\documentclass[graybox]{svmult} 

\usepackage{type1cm}        
\usepackage{makeidx}         
\usepackage{graphicx}        
\usepackage{multicol}        
\usepackage[bottom]{footmisc}

\usepackage{newtxtext}       %
\usepackage[varvw]{newtxmath}       

\usepackage{listings} 
\usepackage{appendix}

\usepackage{multirow,rotating}

\makeindex             

\definecolor{coloresfondo}{rgb}{0.95,0.95,0.95}

\renewcommand{\u}{\varphi}
\renewcommand{\a}{\bar\alpha}
\newcommand{\R}{\mathbb R}

\newcommand{\zmax}{z_{\textup{max}}}
\newcommand{\zmin}{z_{\textup{min}}}
\newcommand{\Gv}{\mathbf v}
\newcommand{\Gw}{\mathbf w}

\newcommand{\Grad}{\mathbf \nabla}
\newcommand{\Gg}{\mathbf {G}}
\newcommand{\Gn}{\mathbf {N}}

\renewcommand{\qed}{\hfill\blacksquare}

\begin{document}

\title*{An Overview of Some Mathematical Techniques and Problems Linking 3D Vision to 3D Printing}
\titlerunning{An Overview of Some Mathematical Techniques and Problems Linking 3DV to 3DP} 
\author{Emiliano Cristiani, Maurizio Falcone, and Silvia Tozza}
\institute{Emiliano Cristiani \at Istituto per le Applicazioni del Calcolo "M. Picone", Consiglio Nazionale delle Ricerche, Via dei Taurini 19,
00185 Rome, Italy, \email{e.cristiani@iac.cnr.it}
\and Maurizio Falcone \at Dept. of Mathematics, Sapienza Università di Roma, P.le Aldo Moro 2, 00185 Rome, Italy  \email{falcone@mat.uniroma1.it}
\and Silvia Tozza \at Dept. of Mathematics, Alma Mater Studiorum Università di Bologna, Piazza di Porta S. Donato 5, 40126 Bologna, Italy \email{silvia.tozza@unibo.it}}
%
%
\maketitle

\abstract*{
Computer Vision and 3D printing have rapidly evolved in the last ten years but interactions among them have been very limited so far, despite the fact that they share some mathematical techniques. We try to fill the gap presenting an overview of some techniques for Shape-from-Shading problems as well as for 3D printing with an emphasis on the approaches based on nonlinear partial differential equations and optimization.  We also sketch possible couplings to complete the process of object manufacturing starting from one or more images of the object and ending with its final 3D print. We will give some practical examples of this procedure.}

\abstract{
Computer Vision and 3D printing have rapidly evolved in the last ten years but interactions among them have been very limited so far, despite the fact that they share several mathematical techniques. We try to fill the gap presenting an overview of some techniques for Shape-from-Shading problems as well as for 3D printing with an emphasis on the approaches based on nonlinear partial differential equations and optimization.  We also sketch possible couplings to complete the process of object manufacturing starting from one or more images of the object and ending with its final 3D print. We will give some practical examples of this procedure.}


\section{Introduction}\label{sec:intro}
In general, the Shape-from-Shading (SfS) problem is described by the so-called \emph{irradiance equation}
\begin{equation}\label{general_irradiance_eq}
	I(x,y) = R(\mathbf{N}(x,y), \mathbf{\Lambda}).
\end{equation}
Assuming the surface is represented by a graph $z=u(x,y)$, for every point $(x,y)$ in the plane, that equation gives the link between the normal  $\mathbf{N}$ to the surface, some parameters related to the light (that we simply denote by a vector $\mathbf{\Lambda}$) and the final image $I$. Note that this is a continuous representation of the image where every pixel is associated to a real value  between 0 and 1 (the classical convention is to associate 0 to black and 1 to white). In practice, the points are discrete with integer coordinates and the values of $I$ have discrete values between 0 and 255 for a graylevel image (the extension to color images is straightforward via the three channels RGB).
Many authors have contributed to 3D reconstruction following the pioneering work by Horn \cite{Horn1970, Horn1986}. Many techniques have been proposed by researchers in computer science, engineering and mathematics, the interested reader can find in the surveys \cite{Durou2020, Durou2008} a long list of references mainly focused on variational methods and nonlinear partial differential equations (PDEs). 

In this chapter we focus on the problem to reconstruct the depth of a surface starting from one or more gray-value images, so that the datum is the shading contained in the input image(s). 
It should  be mentioned that considering different input data, we can describe various problems belonging to the same class of  Shape-from-X problems. These problems share the same goal (recover the 3D shape of the object(s)) and use a single viewpoint. Among these are the Shape-from-Polarization problem \cite{Wolff1991constraining}, the Shape-from-Template problem \cite{Bartoli2015}, the Shape-from-Texture problem \cite{Witkin1981}, the Shape-from-Defocus problem \cite{Favaro2005}, etc.

We concentrate our presentation on the methods based on first order nonlinear PDEs, and in particular eikonal type equations, since these equations also appear in the solution of some problems related to 3D printing, like, e.g., overhang construction and optimization. In this way we try to offer a unified approach that starts from one or more images, allows for the reconstruction of the surface and finally brings to 3D printing of the solid object. Note that in this approach the theoretical framework is given by the theory of viscosity solutions for Hamilton-Jacobi equations, see \cite{Barles1994} and the references therein.

The main focus here is on the modeling of some problems coming from computer vision and 3D printing in a unified framework, for this reason we just sketch some numerical approximation schemes. The interested reader can find in the books \cite{Falcone2014, osherbook, sethianbook} many information on finite difference and semi-Lagrangian schemes for these problems.

\medskip

The chapter is organized as follows: In Section \ref{sec:SfS_single_image}, we present the standard approach for SfS based on a single image considering different models for the surface reflectance, camera and the light source. They all lead to a Hamilton-Jacobi equation complemented with various boundary conditions. 
Section \ref{sec:modeling_more_images} is devoted to the models based on more input images, the typical example comes either for a couple of images taken from different point of view (stereo vision) or from the same viewpoint but under different light source directions (photometric stereo). 
Since every image correspond to a different equation, we will have a system of equations similar to those examined in the previous section. 
We conclude the first part on 3D reconstruction with Section \ref{sec:numerics_exampleSfS}, which gives some hints on the numerical approximation of one of the setup illustrated in Section \ref{sec:SfS_single_image}. 

From Section \ref{sec:from3DVto3DP} we deal with 3D printing. We begin recalling the level-set method for front propagation problems as a common tool for various problems related to 3D printing.
In Section \ref{sec:overhangs} we focus on the overhang issue, presenting three different methods for attacking this important problem. 
In Section \ref{sec:tanzilli} we address instead the problem of creating the infill structure of the object to be printed, also giving some practical examples. 

We conclude the chapter with final remarks and two appendices, where we give some basic information about STL and G-code, two file formats commonly used in 3D printing.

\section{Modeling with a single input image}\label{sec:SfS_single_image}
The modeling of the SfS problem depends on how we describe the three major elements that contribute to determine an image from the light reflection: the light characteristics, the reflectance properties of the surface and the camera. In fact, we can have various light sources as well as different reflectance properties of the surface and one or more cameras.
So we start examining more in detail these three components. 

\subsection{Modelization of the surface reflectance}
\begin{figure}[h!]
	\begin{center}
		\includegraphics[width=0.9\textwidth]{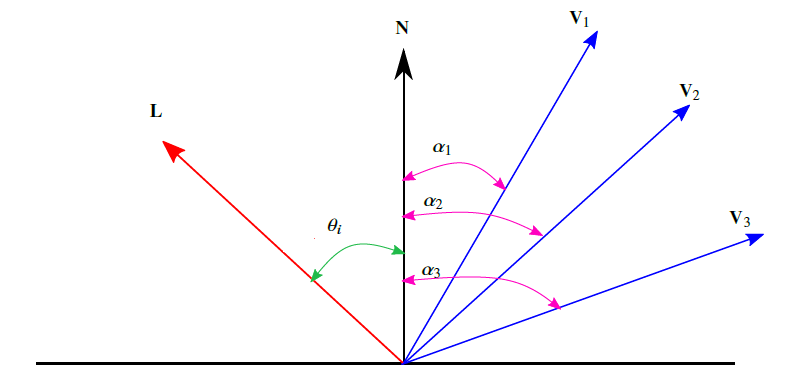}
	\end{center}
	\caption{Lambertian reflectance depends only on the incident angle $\theta_i$. Three different viewers $\mathbf{V}_1$, $\mathbf{V}_2$, and $\mathbf{V}_3$ do not detect any difference in radiance. 
    }
	\label{fig:lambert}
\end{figure}

Depending on how we explicitly describe the function $R$ in Eq. \eqref{general_irradiance_eq}, different reflectance models arise. 
The simplest and more popular model in literature follows the Lambert's cosine law, which establishes that the intensity of a point in the image $I$ in \eqref{general_irradiance_eq} is directly proportional to the cosine of the angle $\theta_i$ between the surface normal and the direction of the incident light. 
The Lambertian model can therefore be represented by the following equation
\begin{equation}\label{eq:lamb_model}
	I(x,y) = \gamma_D \mathbf{N}(x,y) \cdot \mathbf{L} = \gamma_D \cos(\theta_i),
\end{equation}
where $\mathbf{L}$ is a unit vector pointing towards the light source and $\gamma_D$ denotes the diffuse albedo, i.e. the reflective power of a diffuse surface. 
This is a purely diffuse model, since the light is reflected by the surface in all directions, without taking into account the viewer direction, as visible in Fig. \ref{fig:lambert}. 

Another diffuse model is that proposed by Oren and Nayar in the 90s \cite{ON1995, ON1994}. This model is suitable for rough surfaces,  modelled as a set of facets with different slopes, each of them with Lambertian reflectance properties. 
The brightness equation for this model is
\begin{equation}\label{ON_reflectance_model}
	I(x,y)= \cos(\theta _i) (A + B \sin(\alpha) \tan (\beta )\max\{0,\cos(\varphi_r -  \varphi_i)\} ),  
\end{equation}
where $\theta_{i}$ denotes the incident angle, $\theta_{r}$ indicates the angle between the observer  $\mathbf{V}$ and the normal $\mathbf{N}$ directions, $\alpha$ and $\beta$ are defined as $\alpha = \max \left\{ \theta_{i}, \theta_{r} \right\} \hbox{ and } \beta = \min \left\{ \theta_{i}, \theta_{r} \right\}$. 
$\varphi_{i}$ and $\varphi_{r}$ are, respectively, the angles between the projection of the light source direction $\mathbf{L}$,  
or the projection of the viewer direction $\mathbf{V}$, and the $x$ axis onto the $(x,y)$-plane. 
The constants $A$ and $B$ are nonnegative quantities and depend on the roughness parameter $\sigma$ as follows 
\begin{eqnarray*} \label{eq:A_B}
	&& A = 1 - 0.5 \, \sigma^2 (\sigma^2 + 0.33)^{-1} \\
	&& B = 0.45\sigma^2(\sigma^2 + 0.09)^{-1}. 
\end{eqnarray*}

More in general, we can define a reflectance model as sum of three components, which are the ambient, the diffuse and the specular components (cf.\ \cite{Tozza2016}): 
\begin{equation}\label{eq:brightness3components}
	I(x,y) = k_A I_A(x,y) + k_D I_D(x,y) + k_S I_S(x,y),
\end{equation}
with $k_A$, $k_D$, and $k_S$ denoting the percentage of the three components, such that $k_A+k_D+k_S \leq 1$ (if we do not consider absorption phenomena, this sum should be equal to one). 

Following this setting, we can define, for example, the Phong model \cite{Phong1975} as
\begin{equation}\label{eq:PH_brightness}
	I(x,y) = k_A I_A(x,y) + k_D \gamma_D(x,y) (\cos \theta_i) + k_S \gamma_S(x,y) (\cos \theta_s)^{\alpha},
\end{equation}
where the diffuse model is defined as in the Lambertian model, and the third term describes the specular light component as a power of the cosine of the angle $\theta_s$ between the unit vectors $\mathbf{V}$ and $\mathbf{R}(x,y)$ with $\mathbf{R}$ indicating the reflection of the light $\mathbf{L}$ on the surface, $\alpha$ represents the characteristics of specular reflection of a material, and $\gamma_S(x,y)$ denotes the specular albedo. 

In 1977 Blinn \cite{Blinn1977} proposed a modification of the Phong model via an intermediate vector $\mathbf{H}$, which bisects the angle between the unit vectors of the viewer $\mathbf{V}$ and the light $\mathbf{L}$. The brightness equation in this case is 
\begin{equation}\label{eq:BP_brightness}
	I(x,y) = k_A I_A(x,y) + k_D \gamma_D(x,y) (\cos \theta_i) + k_S \gamma_S(x,y) (\cos \delta)^{c},
\end{equation}
where  $\delta$ is the angle between  $\mathbf{H}$ and the unit normal $\mathbf{N}$, and $c$ measures the shininess of the surface.  

Several other reflectance models can be mentioned: the Torrance-Sparrow model \cite{Torrance1967theory}, simplified by Healey and Binford \cite{Healey-Binford86}, the Cook-Torrance model \cite{Cook1982}, the Wolff diffuse reflection model for smooth surfaces \cite{Wolff1994}, the Wolff-Oren-Nayar model that works for smooth and rough surfaces \cite{Wolff1998}, the Ward model for surfaces with hybrid reflection \cite{Ward1992}, etc.

\subsection{Some hints on theoretical issues: viscosity solutions and boundary conditions}\label{subsec:weak_solutions}
In the presentation of the previous models we have always assumed that the surface $u$ is sufficiently regular so that the normal to the surface is always defined. Clearly, this is a very restrictive assumption that is not satisfied for real objects represented in our images. A more reasonable assumption would require almost everywhere (a.e.) regularity, as for a pyramid. 
This will guarantee the uniqueness of the normal vector everywhere but some curves corresponding to the singular points of the surface, i.e. these points form the edges of the object. One could accept solutions in an a.e. sense but this is too much since this simple extension can easily produce infinitely many solutions to the same equation even fixing boundary conditions, this is well known even for the simple eikonal equation on an interval
\begin{equation}
|\nabla u(x)|=f(x) \qquad x\in (a,b)
\end{equation}
 with homogeneous boundary conditions 
 \begin{equation}
 u(a)=u(b)=0
 \end{equation}
whenever $f$ vanishes at one point. One way to select a unique solution is to adopt the definition of viscosity solution $v$ introduced by Crandall and Lions in the 80s.
We will not give this definition here, but it can be easily found, e.g. in  \cite{Barles1994}. For the sequel, just keep in mind that a viscosity solution is an a.e. solution with some additional properties, such as being the limit for $\varepsilon$ going to $0$ of the solution $v^\varepsilon$ of a second order problem
\begin{equation}
-\varepsilon u_{xx}+|\nabla u(x)|=f(x) \qquad x\in (a,b)
\end{equation}
 with homogeneous boundary conditions 
 \begin{equation}
 u(a)=u(b)=0
 \end{equation}
 In some cases, the classical definition is not enough to select a unique solution and additional properties have been considered, e.g. the fact that $v$ must be also the maximal solution, i.e. $v\ge w$ for every $w$ that is an a.e. solution.\\
 
 \vspace{0.2cm} 
{\em Boundary conditions}\\
Another crucial point in the theory of viscosity solutions is the treatment of boundary conditions. All the classical boundary conditions can be considered, but they must be understood (and implemented)  in a weak sense. \\
For the SfS problem, we will consider either Dirichlet boundary conditions
\begin{equation}
u(x)=g(x) \qquad x\in\Gamma=\partial \Omega
\end{equation}
where $g=0$ means that the object is standing on a flat background and a generic $g$ means that we know the height of the surface on $\Gamma$ (this is the case for rotational surfaces). \\
For the front propagation problem, the typical boundary condition will be the homogeneous Neumann boundary condition
\begin{equation}
\frac{\partial u(x)}{\partial {\bf N}} =0 \qquad x\in\Gamma=\partial \Omega.
\end{equation}
We refer again to  \cite{Barles1994} for a detailed analysis of these problems and to \cite{Falcone2014, osherbook, sethianbook} for their numerical approximation.

\subsection{Modelization of the camera and the light}
Assuming that the light source is located at infinity, all light vectors are parallel and we can represent the light direction by a constant vector $\mathbf{L} = (l_1,l_2,l_3)$. 
We assume that the light source is above the surface to be reconstructed, so $l_3 > 0$. 

If we assume that the image is obtained by orthographic projection of the scene, then we can define the surface $\mathcal{S}$ as
\begin{equation}\label{eq:surface_orho}
	\mathcal{S} = \{(x,y,u(x,y)) : (x,y) \in \overline{\Omega}\},
\end{equation}
where $\Omega$ denotes the reconstruction domain. 
Assuming that the surface is regular, a normal vector is defined by
\begin{equation}\label{eq:normal_ortho}
	\mathbf{n}(x,y) = (-\nabla u(x,y), 1).
\end{equation}
Note that orthogonal projection model is rather restrictive since it requires that the distance between the camera and the object is larger than 5-6 times the dimension of the object, however  it is enough to explain  the main features of the SfS problem.
 As an example, the Lambertian model \eqref{eq:lamb_model} under orthographic projection with a single light source located at infinity and uniform albedo constantly equal to one, $\gamma_D \equiv 1$, which means that all the incident light is reflected, 
leads to the following partial differential equation (PDE) in the unknown $u$
\begin{equation}\label{eq:Lamb_PDE_ortho}
	I(x,y) = \dfrac{-\nabla u(x,y)\cdot (l_1,l_2) + l_3}{\sqrt{1+ |\nabla u(x,y)|^2}}.
\end{equation} 
For the SfS problem, we look for  a function $u: \overline{\Omega} \rightarrow \mathbb{R}$ satisfying  the equation \eqref{eq:Lamb_PDE_ortho} $\forall (x,y)\in\Omega$, given an image $I(x,y)$ and a light source direction $\mathbf{L}$. \\
In the particular case of a vertical light source, i.e. $\mathbf{L} = (0,0,1)$, the PDE in \eqref{eq:Lamb_PDE_ortho} reduces to the so-called \emph{Eikonal equation}
\begin{equation}\label{eq:eikonal}
	|\nabla u(x,y)| = \sqrt{\dfrac{1}{I(x,y)^2} - 1} \,.
\end{equation}

Assuming that the image is obtained by a  perspective projection of the scene, we need to distinguish different cases depending on the position of the light source. \\
Supposing a light source located at infinity, the scene can be represented by the following surface
\begin{equation}\label{eq:surface_persp_light_inf}
	\mathcal{S} = \{u(x,y)(x,y,-f) : (x,y) \in \overline{\Omega}\},
\end{equation}
where $f\ge 0$ denotes the focal length, i.e.\ the distance between the retinal plane and the optical center. This is the case of a \textquotedblleft pinhole\textquotedblright camera, that is a simple camera without a lens but with a tiny aperture (the so-called pinhole). 
In this case, the unit normal vector will be
\begin{equation}\label{eq:unit_normal_persp_light_inf}
	\mathbf{N}(x,y) = \dfrac{\mathbf{n}(x,y)}{|\mathbf{n}(x,y)|} = \dfrac{(f \nabla u(x,y), u(x,y) + (x,y)\cdot \nabla u(x,y) )}{\sqrt{f^2 |\nabla u(x,y)|^2 + (u(x,y) + (x,y)\cdot \nabla u(x,y))^2}}
\end{equation}
and so the Lambertian model \eqref{eq:lamb_model} with uniform albedo $\gamma_D \equiv 1$  in this case will lead to the following PDE
\begin{equation}\label{eq:Lamb_PDE_persp_light_inf}
	I(x,y) = \dfrac{f \,\,(l_1,l_2) \cdot \nabla u(x,y) + l_3 (u(x,y) + (x,y)\cdot \nabla u(x,y))}{\sqrt{f^2 |\nabla u(x,y)|^2 + (u(x,y) + (x,y)\cdot \nabla u(x,y))^2}}
\end{equation}
Note that assuming the surface visible, i.e. in front of the retinal plane, then the function $u(x,y)\ge 1, \forall (x,y)\in \overline{\Omega}$ \cite{Prados2003math}. Hence, operating the change of variable $v(x,y) = \ln u(x,y)$, the new variable assumes values greater or equal to 0, $v\ge 0$. \\
The new PDE for the variable $v$ of Eq. \eqref{eq:Lamb_PDE_persp_light_inf} becomes
\begin{equation}\label{eq:Lamb_PDE_in_v_persp_light_inf}
	I(x,y) = \dfrac{(f \,\,(l_1,l_2) + l_3 (x,y)) \cdot \nabla v(x,y) + l_3}{\sqrt{f^2 |\nabla v(x,y)|^2 + (1 + (x,y)\cdot \nabla v(x,y))^2}} \, .
\end{equation}

Now, let us still consider a perspective projection, but with a single point light source located at the optical center of the camera.
This case corresponds to a camera equipped with a flash in a dark place. In this case the parametrization of the scene is a little bit different from the previous case, with a surface $\mathcal{S}$ defined as
\begin{equation}\label{eq:surface_persp_light_center}
	\mathcal{S} = \{  \dfrac{f u(x,y)}{\sqrt{f^2 + |(x,y)|^2}} (x,y,-f) : (x,y) \in \overline{\Omega}\}. 
\end{equation}
For such a surface, the unit normal vector is
\begin{equation}\label{eq:unit_normal_persp_light_center}
	\mathbf{N}(x,y) = 
	\dfrac{(f \, \nabla u(x,y) - \frac{f u(x,y)}{|(x,y)|^2 + f^2} \, (x,y), \nabla u(x,y)\cdot (x,y) +  \frac{f u(x,y)}{|(x,y)|^2 + f^2} \, f)}{\sqrt{ f^2|\nabla u(x,y)|^2 + (\nabla u(x,y) \cdot (x,y))^2 + u(x,y)^2 \frac{f^2}{(f^2 + |(x,y)|^2)} }}
\end{equation}
and the unit light vector depends in this case on the point $(x,y)$ of $\mathcal{S}$, that is
\begin{equation*}\label{eq:light_source_persp_center}
	\mathbf{L}(\mathcal{S}(x,y))  = \frac{(-(x,y),f)}{\sqrt{|(x,y)|^2 + f^2}} \, .
\end{equation*} 
Under this setup, the Lambertian model \eqref{eq:lamb_model} with uniform albedo $\gamma_D \equiv 1$ leads to the the following PDE
\begin{equation}\label{eq:Lamb_PDE_persp_light_center}
	I(x,y) \sqrt{\Bigg(\dfrac{|(x,y)|^2 + f^2}{f^2}\Bigg) \Big[ f^2 |\nabla u(x,y)|^2 + ( (x,y)\cdot \nabla u(x,y) )^2\Big] + u(x,y)^2 } - u(x,y) = 0.
\end{equation}
As done before, assuming the surface visible, then 
\begin{equation*}
	u(x,y)\geq \dfrac{|(x,y)|^2 + f^2}{f^2}\geq 1, \forall (x,y)\in \overline{\Omega}. 
\end{equation*} 
By the same change of variable $v(x,y) = \ln u(x,y)$, considering Lambertian reflectance in  this setup we arrive to the following Hamilton-Jacobi (HJ) equation
\begin{equation}\label{eq:Lamb_PDE_in_v_persp_light_center}
	I(x,y) \sqrt{\Bigg(\dfrac{|(x,y)|^2 + f^2}{f^2}\Bigg) \Big[ f^2 |\nabla v(x,y)|^2 + ( (x,y)\cdot \nabla v(x,y) )^2\Big] + 1 } - 1 = 0 .
\end{equation}

Regarding to the last setup (perspective projection with a light source located at the optical center of the camera), some researchers have introduced a light attenuation term in the brightness equation related to the model considered for real-world experiments (see e.g. \cite{AF06,AF07,Prados2005,Prados2004}), and in \cite{Camilli2017} the well-posedness of the perspective Shape-from-Shading problem for several non-Lambertian models in the context of viscosity solutions is provided, thanks to the introduction of this term. 
This factor takes into account the distance between the surface and the light source and, together with the use of non-Lambertian reflectance models, seems to be useful for better describing the real images. 

\section{Modeling with more input images}\label{sec:modeling_more_images}
Unfortunately, the classical SfS with a single input image is, in general, an ill-posed problem in both the classical and the weak sense, due to the well-known concave/convex ambiguity (see the surveys \cite{Durou2008,Zhang1999}). 
It is possible to overcome this issue by adding information, as, e.g., setting the value of the height $u$ at each point of maximum brightness \cite{Lions1993} or choosing as solution the maximal one, which is proven to be unique \cite{Camilli1999}. Under an orthographic projection, the ill-posedness is still present even if we consider non-Lambertian reflectance models \cite{Tozza2016}. Using a perspective projection, in some particular cases an ambiguity is still visible \cite{Breuss2012,Tozza2022}. 

A natural way to add information is considering more than one input image. We will explain in the following two different approaches which allow to get well-posedness. 

\subsection{Photometric Stereo technique}\label{subsec:PS}
The Photometric Stereo SfS (PS-SfS) problem considers more than one input image, taken from the same point of view but with a different light source for each image (see \cite{Durou2020} for a comprehensive introduction on this problem). 
Just to fix the idea, let us consider two pictures of the same surface, whose intensities are $I_1, I_2$, respectively taken from the light sources $\mathbf{L}', \mathbf{L}''$. 
From the mathematical viewpoint, assuming an orthographic projection of the scene, Lambertian reflectance model, and light sources located at infinity in the direction of the two versors $\mathbf{L}', \mathbf{L}''$, this means considering a system of two equations as
\begin{eqnarray}\label{eq:system_ortho_lamb}
	I_1(x,y) = \gamma_D \mathbf{N}(x,y) \cdot \mathbf{L}' \, ,  \nonumber \\
	I_2(x,y) = \gamma_D \mathbf{N}(x,y) \cdot \mathbf{L}'' \, .
\end{eqnarray} 
Following a differential approach, writing $\mathbf{N}(x,y) = \dfrac{\mathbf{n}(x,y)}{|\mathbf{n}(x,y)|}$, with $\mathbf{n}(x,y)$ defined as in \eqref{eq:normal_ortho}, this corresponds to a system of nonlinear PDEs, in which the nonlinear term is common to both equations and comes from the normalization of the normal vector. Isolating this nonlinear term and the diffuse albedo, the system \eqref{eq:system_ortho_lamb} becomes the following hyperbolic linear problem  
\begin{equation}\label{eq:linEquGrad_ortho}
	\left\{
	\begin{aligned}
		\displaystyle \mathbf{b}(x,y)\cdot\nabla u(x,y)=f(x,y), & \quad\hbox{a.e. $(x,y)\in\Omega$,} \\
		\displaystyle u(x,y)=g(x,y), & \quad\hbox{$\forall(x,y)\in\partial\Omega$,}
	\end{aligned}
	\right.
\end{equation}
to which we added Dirichlet boundary conditions, and where
\begin{equation}\label{bDefinition}
	\mathbf{b}(x,y)=(I_2(x,y)l_1'-I_1(x,y)l_1'',I_2(x,y)l_2'-I_1(x,y)l_2'')
\end{equation}
and
\begin{equation}\label{effe}
	f(x,y)=I_2(x,y)l_3'-I_1(x,y)l_3''.
\end{equation}
It can be shown that the problem \eqref{eq:linEquGrad_ortho} is well-posed \cite{Mecca2013uniqueness}, getting over the concave/convex ambiguity, and it is also albedo independent, a powerful property for real applications. Since in real application is difficult to know the correct boundary conditions, in \cite{Mecca2013shape} the authors shown that it is possible to reconstruct uniquely the surface starting from three images taken from three non-coplanar light sources. Knowing a priori symmetric properties of the surface, the number of required images reduces to a single image generated from any not vertical light source for surfaces with four symmetry straight lines \cite{Mecca2013shape}. 
Non-Lambertian reflectance models have been considered for the PS-SfS problem (see e.g. \cite{Georg03,CJ08,ZMLC10}), and well-posedness has been proved also including specular effects, which allow to reduce artifacts  \cite{Tozza2016direct}.

\medskip

The PS-SfS problem, as well as the problems illustrated in the previous section, can also be solved via a non-differential approach, in which the unknown is the outgoing normal vector to the surface, defined in \eqref{eq:normal_ortho}. 
Differently from the previous differential approach, which works globally, 
this second approach is a local one, that works pixel by pixel in order to find the normal vector field of the unknown surface and then it reconstructs the surface $z = u(x,y)$ all over the domain using the gradient field \cite{Durou2009,Durou2007Integration}. 
Integrability constraint must be imposed in order to limit the ambiguities \cite{Onn1990,Yuille1999}.
We refer the interested reader to \cite{Queau2018normal,Queau2018variational} for a survey on normal integration methods. 

\subsection{Multi-View SfS}\label{subsec:StereoVision}

The Multi-View SfS (MV-SfS) problem considers more than one input image taken from several points of view, but under the same light source for all the images. Differently from the PS-SfS problem, here we have to solve the \textquotedblleft matching issue\textquotedblright (also known as \emph{correspondence problem}), since considering a different viewpoint for each image, the same part of the object(s) we want to reconstruct is located in different pixels of the other images. 

To fix the ideas, you can think to the human binocular vision so the relative depth information should be  extracted by examining the relative positions of object(s) in the two images.
Under suitable assumptions, it is possible to achieve a well-posed problem. As an example, we can follow the Chambolle's approach in \cite{Chambolle1994}, which gets the well-posedness in the case of two cameras under the following assumptions:
\begin{enumerate}
	\item[A1] smooth surface ($u \in C^1$);
	\item[A2] no shadows on the surface;
	\item[A3] noise-free images;
	\item[A4] single light source $\mathbf{L}$ located at infinity;
	\item[A5] Lambertian reflectance model;
	\item[A6] pictures taken with a set of parallel \textquotedblleft standard stereo cameras\textquotedblright \ (i.e.\ two ideal cameras, in which the image is obtained by a simple projection through a pointwise lens, whose coordinate systems are parallel to each other and whose focal distance $f$ is the same);
	\item[A7] no occluded zones on the two images;
	\item[A8] surface represented by a function 
	$Z=u(X,Y)$     
	where the plane $(X,Y)$ is parallel to the focal planes of the cameras and $Z$ remains bounded ($Z\le C$). 
	The two lens are in $(t/2,B,C)$ and $(-t/2,B,C)$, where $t>0$ denotes the translation parameter in the horizontal axis. 
	A point in the scene $(X,Y,Z)$, with $Z\leq C$, appears on the images $I_1$ and $I_2$ in $(x_1,y)$ and $(x_2,y)$ where
	\begin{equation}\label{eq:formulas_relation_Img_points_real_coordinates}
		x_1 = f\frac{X-t/2}{C-Z}, \quad x_2 = f\frac{X+t/2}{C-Z}, \quad y=f\frac{Y-B}{C-Z}.
	\end{equation}
\end{enumerate}

Let $\Omega_1$ be the set of all the points of the first image that appear also on the second one and let us assume that it is a connected set. It should be possible to reconstruct the surface corresponding to the set $\Omega_1$  without ambiguity, if the direction of the light source $\mathbf{L}$ is not too close to the direction $(X,Y)$ of the focal planes, and if the disparity function, which is given by $d:= f\frac{t}{u}$,  where $t$ is the baseline length, 
is already known on the boundary of the set $\Omega_1$.

In one dimension, the problem amounts to look for a disparity function $d(x_1): \Omega_1 \rightarrow \mathbb{R}$ verifying the matching condition
\begin{equation}\label{eq:matching_1D}
	\forall x_1\in\Omega_1, \qquad I_1(x_1) = I_2(x_1+d(x_1)).
\end{equation}
Let us define the set
$$A = \{ a\in\Omega_1 | I_1 \text{ is constant in a neighborhood of } a\}.$$ 
Chambolle has proved  (\cite{Chambolle1994}, Theorem 1) that Eq. \eqref{eq:matching_1D},  associated with the knowledge of $\Omega_1$ and the value of $d$ at one point $a_0\in\overline{\Omega_1}$, is  enough to recover the disparity everywhere where $I_1$ is not locally constant. 
Note that Eq.\ \eqref{eq:matching_1D} gives no information on the value of the disparity in $A$.\\

Once $\Omega_1$ and the disparity $d$ are known, the part of the surface corresponding to the set $\Omega_1$ can be recovered inverting the formulas in \eqref{eq:formulas_relation_Img_points_real_coordinates}, obtaining a parametrized surface and then a normal vector to the surface according to Assumption A1.
By the change of variable $v=\log|d|$, and taking into account Assumption A5, in 1D the SfS equation is
\begin{equation}\label{eq:SfS_1D}
	I_1 = \mathbf{L}\cdot \mathbf{N}
\end{equation}
where $\mathbf{L}=(l_1,l_2)$, with $l_2 < 0$, $\mathbf{N}=(N_1,N_2) = \frac{\mathbf{n}}{|\mathbf{n}|}$. By Assumption A8, we need to have $N_2 < 0$ and thus $N_2 = - \sqrt{1-N_1^2}$, so that
\begin{equation}\label{eq:N_1D}
	\mathbf{N} = (N_1,N_2) = (l_1 I_1 \pm l_2 \sqrt{1-I_1^2}, -\sqrt{1-N_1^2}).
\end{equation}
Hence, there are two possible values for $\mathbf{N}$, except when $I_1 = 1$. Thanks to Eq. \eqref{eq:matching_1D}, the disparity $d$ is uniquely determined in $\Omega_1 \setminus A$. 
In $1D$, A is union of open segments $K = (x_0,x_1)$, on which $I_1$ is constant and \eqref{eq:matching_1D} guarantees the uniqueness of $d(x_0)$ and $d(x_1)$, whereas \eqref{eq:N_1D} gives two possible values for the derivative $N'(x)$ on K. 
$\forall x\in K$, if the two values not coincide, only one of the two $N'$ is compatible with the known values $d(x_0)$ and $d(x_1)$.

To ensure that $d$ can be reconstructed in any case without ambiguity on the whole set $\Omega_1$, we have to know, as a boundary condition, its value on $\partial \Omega_1 = \{ \inf(\Omega_1), \sup(\Omega_1)\}$.

For the extension to the 2D case, we refer the interested reader to the Chambolle's paper \cite{Chambolle1994}, and the results on SfS based on viscosity solutions used therein for the uniqueness result \cite{Crandall1987, Lions1982, Lions1993}.

\section{An example of numerical resolution}\label{sec:numerics_exampleSfS}
In this section, we intend to describe a case of numerical resolution for the classical SfS, whose ingredients that play a role for the 3D reconstruction have been illustrated in Section \ref{sec:SfS_single_image}. As an example, let us fix an orthographic projection, so that the surface normal vector is defined as in Eq.\ \eqref{eq:normal_ortho}, and a single light source located at infinity. Let us consider three different reflectance models: the Lambertian model, the Oren-Nayar model, and the Phong one, whose brightness equations are reported in Eqs. \eqref{eq:lamb_model}, \eqref{ON_reflectance_model}, \eqref{eq:PH_brightness}, respectively. 
As explained in \cite{Tozza2016,Tozza2016comparison}, it is possible to write the three different formulations of these models in a unified fixed point problem as follows:
\begin{equation}\label{eq:fixed_point_SfS}
	\mu v(x,y) = T^M(x,y,v,\nabla v), \quad \hbox{for } (x,y)\in \Omega,
\end{equation}
where $M$ denotes the acronyms of the three models, i.e. $M=L, ON, PH$, $v$ is the new variable introduced by the exponential Kru\v zkov change of variable \cite{Kruzkov1975} $\mu v(x,y) = 1- e^{-\mu u(x,y)}$, which is useful for analytical and numerical reasons. In fact, setting the problem in the new variable, $v$  will have  values in $[0, 1/\mu]$ instead of $[0,\infty)$ as the original variable $u$ so an upper bound will be easy to find. 
The parameter $\mu$ is a free constant which does not have a specific physical meaning in the SfS problem, but it can play an  important role also in the convergence proof (see the remark following the end of Theorem 1 in \cite{Tozza2016}). 
The operator $T^M$ can be defined as
\begin{equation}\label{eq:T_operator}
	T^M :=\min\limits_{a\in \partial B_3} \{ {\bf b}^M(x,y,a)\cdot \nabla v(x,y)+ f^M(x,y,z,a,v(x,y))\}
\end{equation}
where the vector field ${\bf b}^M$ and the cost $f^M$ vary according to the specific model and to the case.

This unified formulation gives the big advantage to solve numerically different problems in a unified way. 
In order to obtain the fully discrete approximation, we adopt here the semi-Lagrangian approach described in the book \cite{Falcone2014}, so that we get  
\begin{equation} \label{eq:generic_operator}
	W_i=\widehat T_i^M(\mathbf{W}), 
\end{equation}
where $\mathbf{W}$ is the vector solution giving the approximation of the height $u$ at every node $x_i$ of the grid, and we use the notation $W_i=w(x_i)$, $i$ indicating a multi-index, $i=(i_1,i_2)$. 
Denoting by $G$ the global number of nodes in the grid, the operator corresponding to a general  oblique light source is $\widehat T^M:\R^G\rightarrow \R^G$ that is defined componentwise by
\begin{eqnarray} \label{eq:generic_T}
	\widehat T_i^M(\mathbf{W}) \hspace{-0.5mm} := 
	\min\limits_{a\in\partial B_3} \{ e^{-\mu h} I[\mathbf{W}](x_i^+) - \tau F^M(x_i,z,a) \} + \tau 
\end{eqnarray}
where $I[\mathbf{W}]$ represents an interpolation operator based on the values at the grid nodes, and 
\begin{eqnarray}
	&& x_i^+:= x_i + h b^M(x_i,a) \\
	&&\tau := (1-e^{-\mu \, h})/{\mu} \\
	&&F^M(x_i,z,a) := P^M(x_i,z) a_3 (1-\mu W_i) \\
	&&P^M: \Omega\times\R \rightarrow \R \hbox{ is continuous and nonnegative}.
\end{eqnarray}
Since  $w(x_i + h b^M(x_i,a))$ is approximated via $I[\mathbf{W}]$ by interpolation on $\mathbf{W}$ (which is defined on the grid $G$), it is important to use a monotone interpolation in order to preserve the properties of the continuous operator $T^M$  in the discretization. To this end, the typical choice is to apply a piecewise linear (or bilinear) interpolation operator $I_1 [\mathbf{W}]: \Omega\rightarrow \R$   which allows to define a function defined for every $x\in \Omega$ (and not only on the nodes)
\begin{equation}\label{int1}
	w(x)= I_1[\mathbf{W}](x)=\sum_j \lambda_{ij} (a) W_j
\end{equation}
where
\begin{equation}\label{int2}
	\sum_j \lambda_{ij} (a)=1 \quad \hbox{for} \quad x=\sum_j \lambda_{ij}(a) x_j.
\end{equation}
A simple explanation for \eqref{int1}-\eqref{int2} is that the coefficients $\lambda_{ij}(a)$ represent the local coordinates of the point $x$ with respect to the grid nodes (see \cite{Falcone2014} for more details and other choices of interpolation operators). Clearly, in \eqref{eq:generic_T} we apply the interpolation operator to the point $x^+_i=x_i + h b^M(x_i,a)$ and we denote by $w$ the function defined by $I_1[\mathbf{W}]$.

\begin{figure}[h!]
	\begin{center}
		\includegraphics[width=2.2cm]{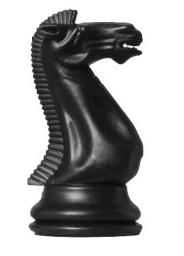} \hspace{2cm}
		\includegraphics[width=2.2cm]{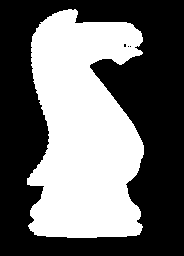}
	\end{center}
	\caption{From left to right: Input image and related mask adopted for the black chess horse test. Pictures taken and adapted from \cite{Tozza2016comparison}.}
	\label{fig:Horse_input_mask}
\end{figure}

\begin{figure}[h!]
	\begin{center}
		\begin{tabular}{ccccccc}
			& Lambertian &
			Oren-Nayar &
			Phong 
			\\ \hline	\\
			\normalsize	
			\begin{sideways} \hspace{0.5cm}{$3D\, SURFACE$} \end{sideways} &
			\includegraphics[width=3.2cm]{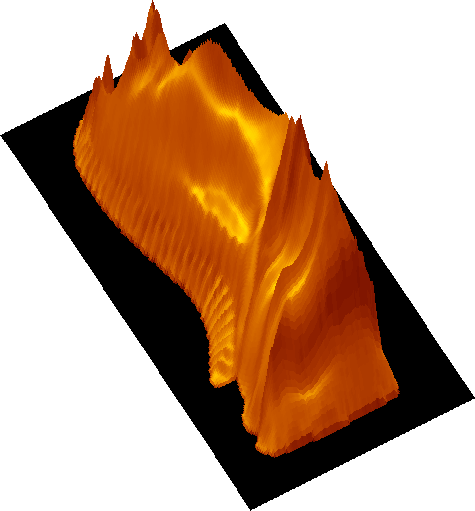} &
			\includegraphics[width=3.2cm]{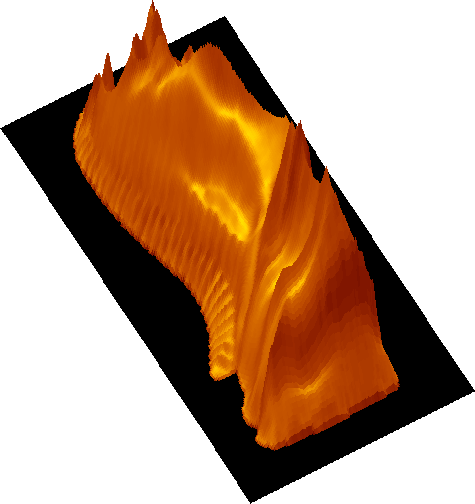} &
			\includegraphics[width=3.2cm]{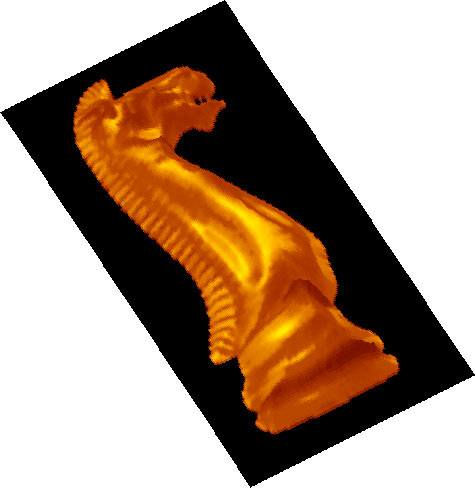} \\ \hline
			\normalsize	
			\\[-0.5cm]
		\end{tabular}
	\end{center}
	\caption{Black chess horse:  3D reconstructions related to the three models with $\sigma = 0.2$ in the Oren-Nayar model, and $k_S = 0.8, \alpha = 1$ for the Phong model. Light source $\mathbf{L} = (0,0,1)$, viewer $\mathbf{V} = (0,0,1)$. Pictures taken and adapted from \cite{Tozza2016comparison}.}
	\label{fig:HorseSet}
\end{figure}

Under suitable assumptions, it is possible to prove that the fixed-point operator $\widehat T^M$ is a contraction mapping in $L^{\infty}([0,1/\mu)^G)$, is monotone, and $0\leq \mathbf{W} \leq \frac{1}{\mu}$ implies $0\leq \widehat T^M(\mathbf{W})\leq \frac{1}{\mu}$. These properties allow to prove a convergence result for the algorithm based on the fixed-point iteration
\begin{equation}\label{def:sucpf}
	\left\{ \begin{array}{ll}
		\mathbf{W}^n &=\widehat T^M(\mathbf{W}^{n-1}),\\
		\mathbf{W}^0 & \textrm{given}.
	\end{array} \right.
\end{equation}
We refer the interested reader to \cite{Tozza2016} for the proofs and more details. 
An example of numerical results by using the semi-Lagrangian scheme illustrated above is visible in Fig. \ref{fig:HorseSet}, where the 3D reconstructions obtained by using the operators related to the three reflectance models are shown. This figure is related to a real image of chess horse used as input image and visible in Fig. \ref{fig:Horse_input_mask} with the corresponding mask adopted. 
For this numerical test, the Phong model performs clearly better, being the surface shiny.

\paragraph{\bf A funny counterexample of non-uniqueness}
Just for fun, we have tried to 3D-print the surface reconstructed from the famous Lena image under the assumptions of orthographic Lambertian model: although the surface mismatches completely the real person, a picture taken from above correctly returns the input image, as expected!  

\begin{center}
	\includegraphics[width=3cm]{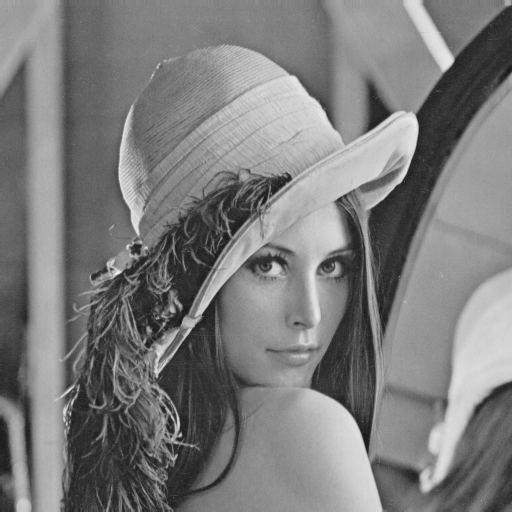}
	\includegraphics[width=3cm]{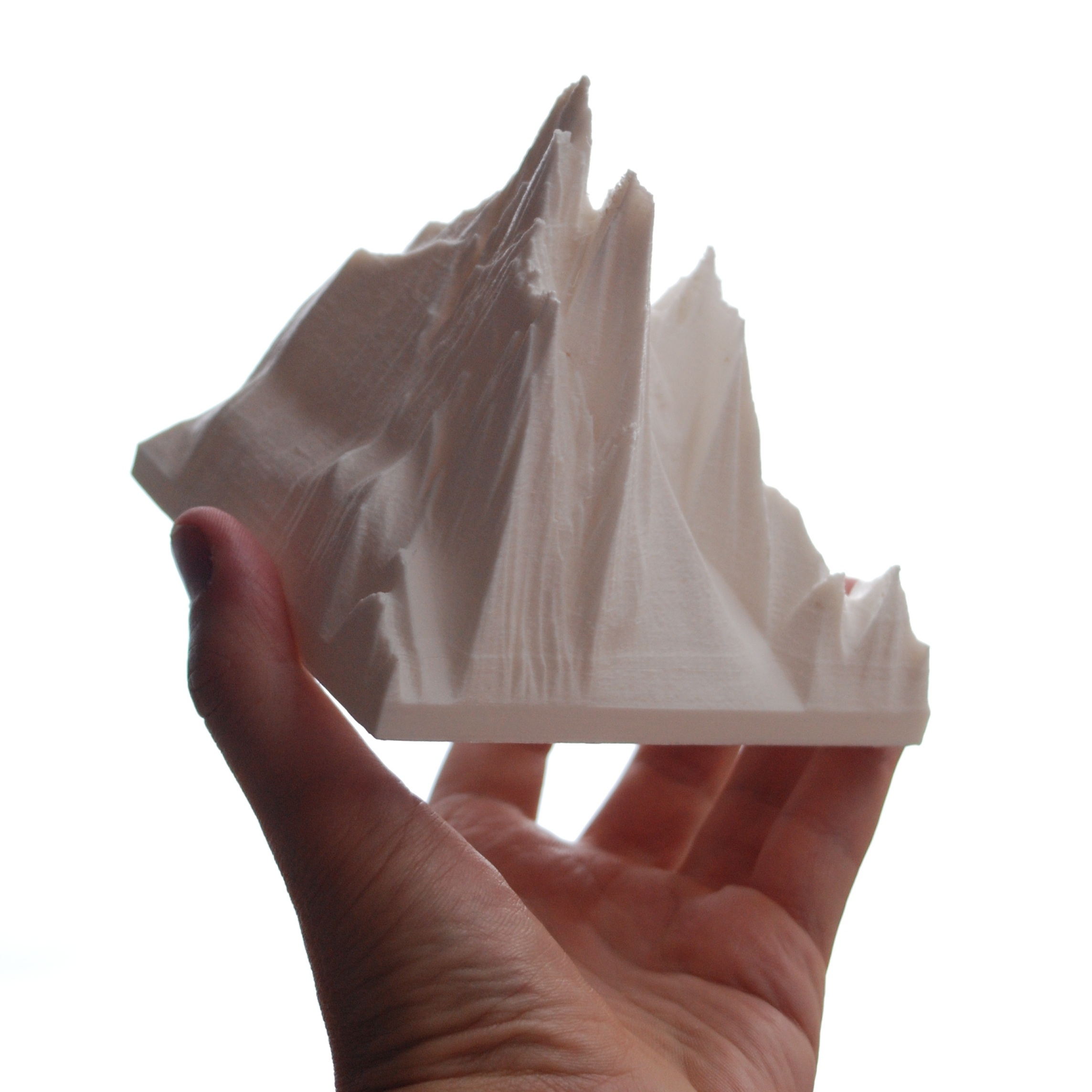}
	\includegraphics[width=3cm]{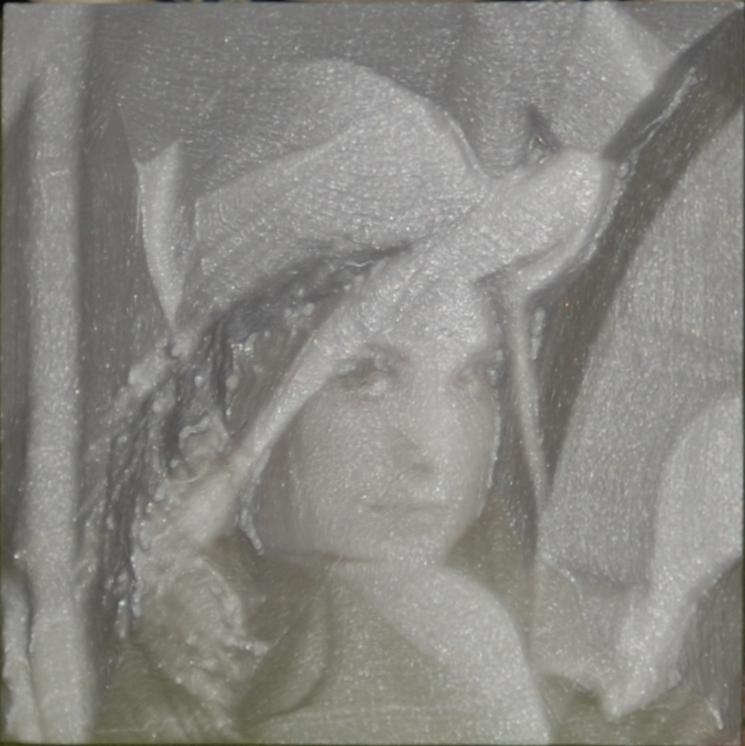}\\
\end{center}

\section{Moving from 3D vision to 3D printing} \label{sec:from3DVto3DP}
Surfaces like the ones depicted in Fig.\ \ref{fig:HorseSet}, which represent the graphs of different functions, can be easily transformed in watertight (closed) surfaces of 3D objects, in order to manufacture them with suitable 3D printers. 
However, when it comes the time to actually 3D-print the object, possibly reconstructed via SfS techniques, many additional issues could arise. In the following we will review some of them, with special emphasis to those which are most related to 3D vision and/or Hamilton-Jacobi equations, level-set method and front propagation problems \cite{osherbook, sethianbook}. 
The related basic mathematical theory will be recalled in Section \ref{sec:LSmethod}.

\subsection{Overview}\label{sec:3DPoverview}
Here we list a brief summary of problems often encountered in 3D printing. For broader overviews, we refer the reader to \cite{el-sayegh2020ACME, livesu2017CGF, panda} and references therein.
\begin{itemize}
	\item \textbf{Partitioning}. Sometimes it is necessary to divide a 3D model into multiple printable pieces, so as to save the space, to reduce the printing time, or to make a large model printable by small printers \cite{attene2015CGF}. This problem was attacked by means of a level-set based approach in \cite{yao2015TOG}. 
	\item \textbf{Overhang, supports, orientation}. 3D printers based on fused deposition modeling (FDM) technology create solid object layer by layer, starting from the lowest one. 
	As a consequence, each layer can only be deposited on top of an existing surface, otherwise the print material falls down and solidifies in a disorderly manner. 
	Little exceptions can be actually handled, i.e.\ the upper layer can protrude over the lower layer within a certain limit angle $\bar\alpha$. Typically $\bar\alpha=\frac{\pi}{4}$, because of the so-called $45$ degree rule, though it actually depends on the 3D printer settings, print material, cooling, etc.
	In recent years many solutions were proposed to deal with this problem: 
	first of all, the object can be rotated to find an orientation, if any, which does not show overhangs \cite{ezair2015CeG}. 
	Another widely adopted solution consists in adding \emph{scaffolds} during construction, in order to support the overhanging regions. In this respect, it is important to note that scaffolds represents an additional source of printed material and printing time, and have to be manually removed at the end of the process. This is cumbersome, time-consuming and reduces the quality of the final product, see e.g., \cite{cacace2017AMM} and references therein.
	Finally, overhangs issues can be fixed \emph{in the design phase of the object}, running a shape optimization algorithm which modifies the unprintable object in order to make it printable while preserving specific features \cite{allaire2018SMO, allaire2017AMNS, allaire2017JCP, allaire2017CRASP, langelaar2018SMO}.\\  
	In the following sections we recall three approaches based on nonlinear PDEs. All of them recast the problem of overhangs in a front propagation problem and then use the level-set method to solve the problem.
	\item \textbf{Infill.} Printing fully solid objects is often not convenient because of the large quantity of material to be used. The problem reduces to finding the optimal inner structure for saving material while keeping the desired rigidity and printable features. 
	In Section \ref{sec:tanzilli} we will detail a method inspired by \cite{dhanik2010IJAMT,kimmel1993CAD} and based on the front propagation problem to compute ad hoc infill patterns.
	\item \textbf{Appearance.} A problem about 3D printing which is strongly related to 3D vision and Shape-from-Shading problem regards the way the objects appear, i.e.\ how they reflect light when illuminated. 
	This problem is very important when the goal is to replicate (multi-material) real objects with complex reflectance features using a single printing material, possibly with simple Lambertian reflectance. 
	The idea is to print the object with small-scale ripples on the surface in order to control the reflectance properties for a given direction (or all directions) of the light source. 
	For example, the approach proposed in \cite{pellacini2013TOG} allows one to create a shape which reproduces the appearance of costly materials (e.g., velvet) using cheaper ones, duly orienting small facets on the surface and then covering them by ad hoc varnish. In this way it is possible to reproduce a given target Bidirectional Reflectance Distribution Function (BRDF) \cite{ngan2005proc}. Note that this is the same ingredient used in Shape-from-Shading for the opposite problem, i.e.\ reconstruct the shape from reflectance features.
	The possibility of controlling the reflectance properties of an object by means of its physical shape opens many interesting possibilities, one can refer to \cite{piovarci2017, rouiller2013CGA, weyrich2009TOG} for some examples.  
	\item \textbf{Slicing \& toolpath generation.} Once the object to be printed is defined, typically in terms of triangles covering its surface, one has to cut it with a sequence of parallel planes in order to compute the \emph{layers} used in the printing process. This procedure is not trivial in case of very complicated (self-intersecting, overlapped) objects. 
	Once the layers are created, the exact trajectory of the nozzle must be defined in order to cover all the layer points; see, among others, \cite{jiang2020MM, jin2014AM}.
	Interestingly, this problem can be seen as a generalized traveling salesman problem. 
\end{itemize}

In the following we will review some mathematical approaches for the overhang and the infill problems, all based on the level-set method for front propagation problems. 
Before that, we recall some basic mathematical results for the reader's convenience.

\subsection{Front propagation problem, level-set method and the eikonal equation}\label{sec:LSmethod}
The level-set method was introduced in \cite{osher1988JCP}  to track the evolution of a $(d-1)$-dimensional surface (the front) embedded in $\R^d$ evolving according to a given velocity vector field $\Gv:\R^+\times\R^d\to\R^d$. 
Let us briefly recall the method in the case of $d=3$.
It is given a bounded closed surface $\Sigma_0:U\subset\R^2\to\R^3$ which represents the front at initial time $t=0$. 
We denote by $\Sigma_t$ its (unknown) evolution under the action of $\Gv$ at time $t$ and by $\Omega_t$ the 3D domain strictly contained in $\Sigma_t$ so that $\Sigma_t = \partial \Omega_t$, for all $t\geq 0$.

The main idea of the level-set method consists in increasing the space dimension of the problem by 1, in order to represent the 3D surface as a level-set of a 4D function 
$\u(t,x,y,z):\R^+\times\R^3\to\R$. 
This function is defined in such a way that
\begin{equation}
	\Sigma_t=\{(x,y,z)\ :\ \u(t,x,y,z)=0\},\quad \forall \, t\geq 0.
\end{equation}
\noindent In this way the surface is recovered as the zero level-set of $\u$ at any time. 
As initial condition for $\u$, one chooses a function which changes sign at the interface,
\begin{equation}\label{proprietaLS}
	\u(0,x,y,z)
	\left\{
	\begin{array}{ll}
		>0, & \textrm{if } (x,y,z)\notin \overline{\Omega_0}, \\
		=0, & \textrm{if } (x,y,z)\in \Sigma_0, \\
		<0, & \textrm{if } (x,y,z)\in \Omega_0.
	\end{array}
	\right.
\end{equation}
A typical choice for $\u(0,x,y,z)$ is the signed distance function from $\Sigma_0$, although is it not smooth and can lead to numerical issues. 
It can be proved \cite{sethianbook} that the level-set function $\u$ at any later time satisfies the following Hamilton--Jacobi equation
\begin{equation}\label{LSequation}
	\partial_t \u(t,x,y,z)+\Gv(t,x,y,z)\cdot \Grad\u(t,x,y,z)=0,\qquad t\in\R^+,\ (x,y,z)\in\R^3,
\end{equation}
with an initial condition $\u(0,x,y,z)=\u_0(x,y,z)$ satisfying \eqref{proprietaLS}. Here $\Grad=(\partial_x,\partial_y,\partial_z)$ denotes the gradient with respect to the space variables only. 

Several geometrical properties of the evolving surface can be still described in the new setting, again by means of its level-set function $\u$. 
For example, it possible to write the unit exterior normal $\Gn$ and the (mean) curvature $\kappa$ in terms of $\u$ and its derivatives. More precisely, we have
\begin{equation}\label{normal=gradient}
	\Gn=\frac{\Grad\u}{|\Grad\u|} \quad \text{ and } \quad \kappa= \Grad \cdot \Gn.
\end{equation}

If the evolution of the surface occurs in the normal direction of the surface itself, i.e.\ the vector field has the form $\Gv = v\Gn$ for some scalar function $v:\R^+\times\R^3\to\R$, the equation \eqref{LSequation} turns into the evolutive eikonal equation
\begin{equation}\label{eikonaleevolutiva}
	\partial_t \u(t,x,y,z)+v(t,x,y,z)|\Grad\u(t,x,y,z)|=0. 
\end{equation}
Moreover, if the scalar velocity field $v$ also depends on the direction of propagation $\Gn$, using \eqref{normal=gradient} we get the \emph{anisotropic eikonal equation}
\begin{equation}\label{eikonaleevolutivaaniso}
	\partial_t \u(t,x,y,z)+v\left(t,x,y,z,\frac{\Grad\u}{|\nabla\u|}\right)|\Grad\u(t,x,y,z)|=0.
\end{equation}

Finally, in the particular case of a constant-sign time-independent velocity field $v(x,y,z,\Gn)>0$ or $v(x,y,z,\Gn)<0$ for all $(x,y,z)$ we are guaranteed that the surface evolution is \emph{monotone}, i.e.\ the surface is either enlarging or shrinking for all times $t$. 
In this case, Eq.\ \eqref{eikonaleevolutivaaniso} can be written in the following stationary form
\begin{equation}\label{eikonalestazionariaaniso}
	v\left(x,y,z,\frac{\nabla T}{|\nabla T|}\right)|\Grad T(x,y,z)|=1 
\end{equation}
and the surface $\Sigma$ can be recovered from $T$ as
$$
\Sigma_t=\{(x,y,z) \ : \ T(x,y,z)=t\}, \quad \forall t\geq 0.
$$

All these equations are particular Hamilton-Jacobi equations for which many theoretical results and numerical methods were developed in the last years. 
We refer the reader to \cite{Falcone2014} for details.

\subsection{Computation of the signed distance function from a surface}\label{sec:LS.df}
The computation of the signed distance function $\u_0$ from $\Sigma_0$ is a problem \emph{per se}. 
Speaking of 3D objects to be 3D-printed, we can assume that the surface $\Sigma_0$ is watertight and that it is given by means of a triangulation (typically in the form of a .STL file, see Appendix A). 
Each triangle (facet) $f$ is characterized by the 3D coordinates of its three vertices. 
Moreover, vertices are oriented in order to distinguish the internal and the external side of the facet (right-hand rule).

Given a point $(x,y,z)\in\R^3$, it is easy to find the distance $d\big((x,y,z),f\big)$ between the point and the facet, so that the unsigned distance from the surface is simply given by
$$
d((x,y,z),\Sigma_0)=\min_f d((x,y,z),f).
$$ 
The computation of the distance's sign is instead more tricky since one has to check if the point is internal or external to the surface. 
One method to solve the problem relies on the fact that the solid angle subtended by the whole surface at a given point is maximal and equal to $4\pi$ if and only if the point is internal. 
Therefore, one can sum all the solid angles subtended by the facets at the point and check if it equals $4\pi$ or not. In the first case the point is internal to the surface, in the second case it is external.

Note that the solid angle itself should be \emph{signed}, in the sense that it must be positive if the point looks at the internal part of the facet, negative otherwise. A nice algorithm to compute the signed solid angle between a point and a triangle was given by van Oosterom and Strackee \cite{oosterom1983}.

\section{Handling overhangs}\label{sec:overhangs}
In this section we deal with the problem of overhangs, see Section \ref{sec:3DPoverview}. 
We recall three approaches recently proposed in the context of Applied Mathematics. All of them recast the problem of detecting/resolving overhangs in a \emph{front propagation problem}.

\subsection{Detecting overhangs via front propagation}
Here we recall the method proposed by van de Ven et al.\ in \cite{vandeven2018proc} and then extended by the same authors in \cite{vandeven2018SMO, vandeven2020CMAME, vandeven2021SMO}.
The authors start noting the resemblance between the front propagation problem, see Section \ref{sec:LSmethod}, and the additive manufacturing process where with every added layer, the lower boundary of the product advances.
Therefore, the first layer, i.e.\ the layer leaning on the build plate, is considered as the initial front and it is propagated by solving the anisotropic stationary eikonal equation \eqref{eikonalestazionariaaniso}.

Overhangs are detected by solving the equation with two different velocity fields.
In the first case, the velocity field is chosen is such a way that the front reaches, at the same time, all points of the object which are at the same height.
In this case it is not actually needed to solve the equation since the arrival time is a measure of distance to the base plate,
the minimum arrival time is simply given by
\begin{equation}
	T_1(x,y,z)=\frac{(x,y,z)\cdot \mathbf h}{v_0}
\end{equation}
where $\mathbf h$ is a unit vector pointing in the build direction (assuming that the origin of the coordinate system is on the build plate), and $v_0$ is the propagation speed, which can be interpreted as the printing rate.

In the second case, which is the crucial one, the velocity field is chosen to be constant and equal to $v_0$ (thus recovering the classical isotropic eikonal equation) except when the front travels in a direction lower than the minimum allowable overhang angle. 
When this happens, the velocity is \emph{decreased}, so that the arrival time will be larger than $T_1$ at the same point.
In \cite{vandeven2020CMAME} this effect is obtained defining
\begin{equation}
	v(\mathbf a;\bar\alpha)=\frac{v_0}{\max\{\tan(\bar\alpha)\|\mathbf P\mathbf a\|,|\mathbf h\cdot \mathbf a|\}}
\end{equation}
where $\mathbf a=\frac{\nabla T}{|\nabla T|}$ is the direction of propagation, 
$\bar\alpha$ is a given overhang angle (parameter), and  
$\mathbf P$ is the projection on the plane orthogonal to $\mathbf h$, defined as 
$\mathbf P = \mathbf I - \mathbf h \otimes  \mathbf h$, with $\otimes$ denoting the outer
product. See \cite{vandeven2020CMAME} for a detailed motivation of this choice.

Finally, a straightforward comparison of the two solutions reveals the overhanging regions, associated to different arrival times of the front.

\subsection{Fixing overhangs via level-set method 1: a direct approach}\label{rocchi}
Here we recall the method proposed by Cacace et al.\ in \cite{cacace2017AMM}.
The idea is to recast the problem in the level-set framework considering the surface $\Sigma$ of the object $\Omega$ to be 3D-printed as an evolving front. If the object is unprintable without scaffolds, its shape is modified letting its surface evolve under an \emph{ad hoc} vector field $\Gv$ to be suitably defined, until it becomes fully printable, i.e.\ it has no hanging parts exceeding the limit angle $\alpha$. 
The new object $\Omega_*$ is then actually printed and the difference $\Omega_*\backslash\Omega_0$ is finally removed. 
Note that the difference $\Omega_*\backslash\Omega_0$ can be easily identified by standard techniques and consequently printed with a different material (e.g., a soluble filament) or with a different printing resolution.

\begin{figure}[h!]
	\centering
	\begin{tabular}{cc}
		\includegraphics[scale = 0.12]{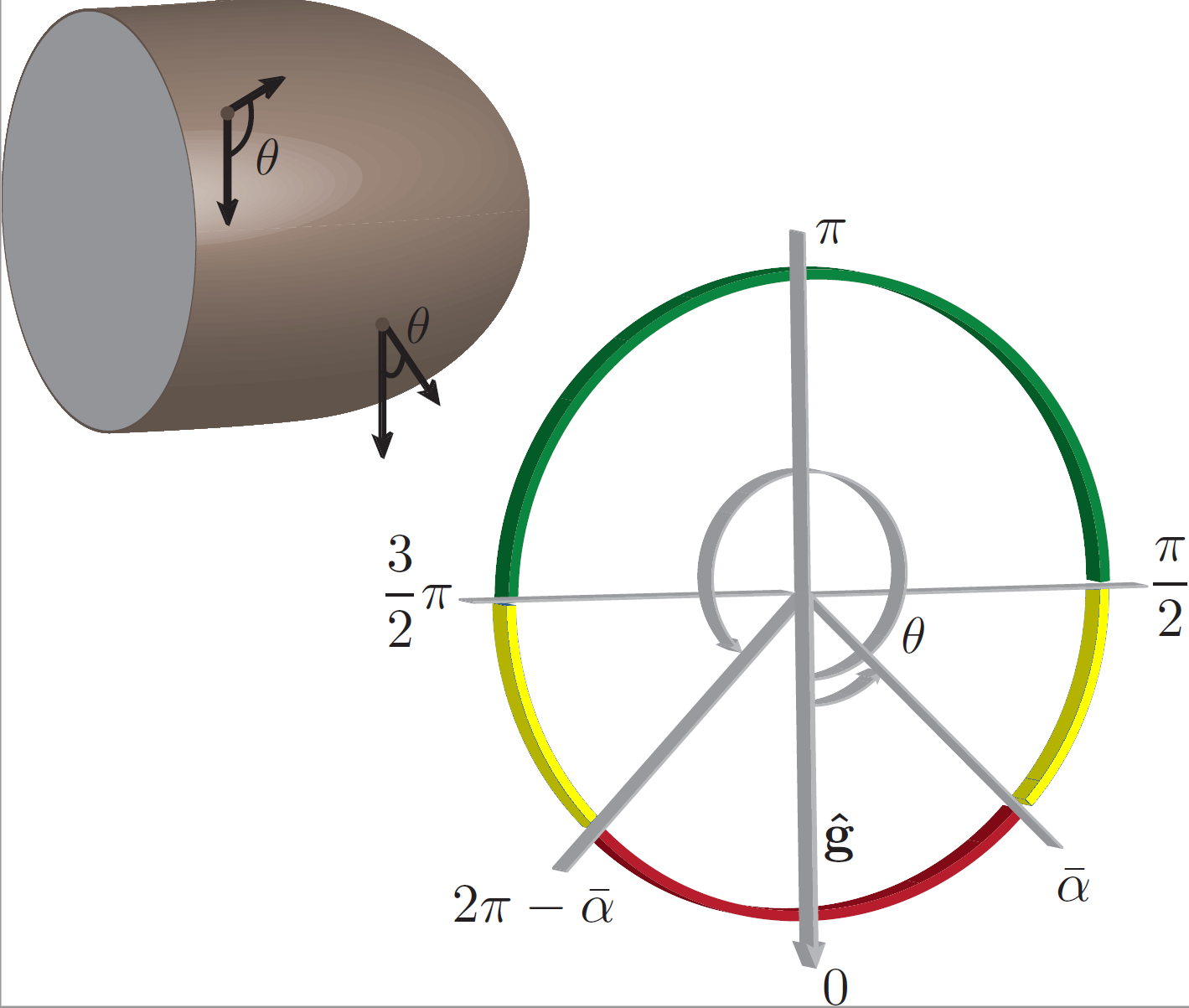} \qquad \qquad &
		\includegraphics[scale = 0.13]{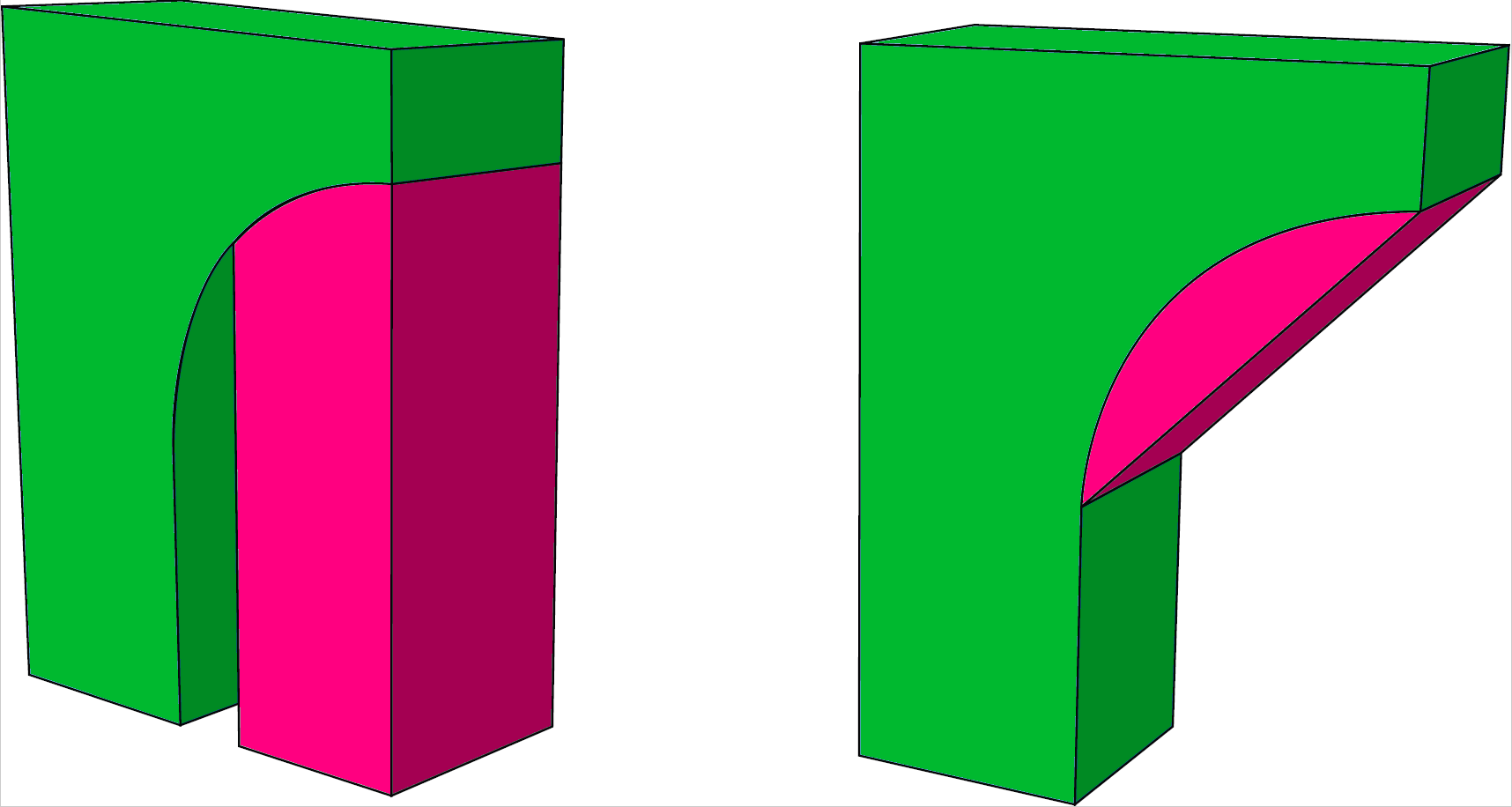} \\
		\footnotesize (a) & \footnotesize (b) 
	\end{tabular}
	\caption{\footnotesize{(a) Unprintable (red), modifiable (yellow) and safe (green) points with respect to the counter-clockwise angle $\theta$ between the gravity $\Gg$ and the normal $\Gn$. Modifiable and safe points are printable. (b) The left grey support wastes a lot of material contrary to the chamfer on the right that saves more material and keeps the printability of the overhang as well.}}
	\label{theta_overhang}
\end{figure}

To this end, the surface $\Sigma$ of the object $\Omega$ is divided into three subsets, on the basis of their \emph{printability}. 
Denote by $\Gg=(0,0,-1)$ the unit gravity vector, and again by $\Gn(x,y,z)$ the exterior unit normal to the surface $\Sigma$ of the object $\Omega$ at the point $(x,y,z)$. 
Moreover, let
\begin{equation}
	\theta(\Gn) := \arccos \big(\Gg\cdot\Gn\big) 
\end{equation}
be the angle between $\Gg$ and $\Gn$. 
\begin{definition}
	\label{def:printable}
	\emph{A point $(x,y,z)$ of the surface $\Sigma$ is said to be}
	\begin{center}
		\begin{tabular}{rl}
			\textit{unprintable,}  & if $\theta\in[0,\a)\cup(2\pi-\a,2\pi]$, \\
			\textit{safe,}                & if $\theta\in[\pi/2,3\pi/2]$, \\
			\textit{modifiable,}   & otherwise,
		\end{tabular}
	\end{center}
	\noindent where $\a$ is the given limit angle, see Fig.\ \ref{theta_overhang}(a).
\end{definition}  

While the first two definitions are immediately clear, it is worth to spend some words on the third one. Modifiable points are indeed printable since the overhang is sufficiently small. On the other hand, it could be convenient to move those points as well in order to make printable the unprintable ones. This guarantees a sufficient flexibility to shape the object conveniently and not to create long supports like the one depicted on the left of Fig.\ \ref{theta_overhang}(b). Definition \ref{def:printable} can be extended by saying that the set of both modifiable and safe point constitute the overall printable points. 

\medskip

We now focus on the construction of the vector field $\Gv$. 
Recalling equation \eqref{eikonaleevolutiva}, we choose $\Gv=v\Gn$ for some scalar function $v$, possibly depending on $\Gn$ and $\kappa$ (but not on $t$). 

In the following we denote by 
$$P(\omega):=\omega^+ \quad \text{and} \quad  M(\omega):=\omega^-, \qquad \omega\in\R,$$ 
the positive and negative part, respectively.

\medskip

\noindent \textit{Positivity and build plate.} It is needed that $\Omega_0\subseteq \Omega_t\subseteq\Omega_*$, for all $t\geq 0$, since once the object is printed the material can be removed but not added. 
This is why $v \geq 0$ is a necessary assumption, i.e., 
the movement of each point of the surface $\Sigma$ has to be along the normal exterior direction $\Gn$. Furthermore, the object cannot move under the build plate, supposed at a fixed height $z=\zmin \in \R$. Therefore it is imposed that $v=0$ if $z \leq \zmin$.

\medskip

\noindent \textit{Movement of unprintable points.} 
We introduce the term
\begin{equation}
	v_1(\Gn;\a):=P(\cos\theta(\Gn)-\cos\a),
\end{equation}
which lets the unprintable points move outward. The speed is higher whenever $\theta$ is close to $0$, which represents the (hardest) case of a horizontal hanging part.

\medskip

\noindent \textit{Rotation.} It is convenient introducing a rotational effect in the evolution which avoids the unprintable regions to evolve ``as it is'' until they touch the build plate. 
To this end the term $v_1$ is multiplied by $(\zmax-z)$, where $\zmax \in \R$ is the maximal height reached by the object. This term simply increases the speed of lower points with respect to higher ones. This makes the lower parts be resolved (or eventually touch the built plate) before the higher parts, thus saving material. 

\medskip

\noindent \textit{Movement of modifiable points.} Modifiable points are moved, if necessary, by means of the following term in the vector field
\begin{equation}
	v_2(\kappa) := M(\kappa).
\end{equation}
It moves outward the points with negative curvature until it vanishes, i.e.\ the surface is locally flat. In particular, it moves concave corners and let modifiable points become a suitable support for the still unprintable points above.

\medskip

\noindent \textit{Blockage of safe points.} Finally, it is necessary to exclude from the evolution the safe points of the object. In order to identify them, we use the sign of the third component $n_3$ of the unit exterior normal vector $\Gn$. 

\medskip

\noindent \textit{Final model.} 
By putting together all the terms one obtains 
\begin{multline}\label{complete_v}
	v(x,y,z,\Gn,\kappa;\a) :=\\ 
	\left\{
	\begin{array}{ll}
		C_1\, (\zmax-z)v_1(\Gn;\a) + C_2\,v_2(\kappa), & \text{ if } n_3<0 \text{ and } z > \zmin,\\
		0, & \text{ otherwise, }
	\end{array}
	\right.
\end{multline}
with $C_1, C_2>0$ positive constants (model parameters). 
The result expected from a such vector field is an evolution similar to the one depicted on the right in Fig.\ \ref{theta_overhang}(b), corresponding to a support whereby the angle $\theta$ in each of its point is less or equal to $\a$.

The surface evolution must be stopped at some final time $t_{f}>0$. 
%
It is convenient to check directly (at every time $t<t_f$) whether the overall surface is printable or not, according to Definition \ref{def:printable}, rather than waiting that the velocity field vanishes completely. 
More precisely, the evolution is stopped when all the points belonging to the zero level-set are safe or modifiable, i.e., printable.

By construction, the surface always evolves towards a printable object. Indeed,
any non-printable part of the surface is forced to move downward, and the surface has to stop once the build plate is reached. 
Nevertheless, we have no guarantee that the final object is ``optimal'' in terms of additional printing material. 
In the worst-case scenario the surface evolves until it touches the build plate, obtaining something similar to the results
depicted on the left in Fig.\ \ref{theta_overhang}(b). 
Nevertheless, the method works fine in most cases, see \cite{cacace2017AMM} for some numerical results.
\subsection{Fixing overhangs via level-set method 2:  topological optimization with shape derivatives}
Here we present very briefly the method proposed by Allaire et al.\ in \cite{allaire2017JCP} (see also the related papers \cite{allaire2017AMNS, allaire2017CRASP} and \cite{allaire2020SMO, allaire2018SMO, allaire2018M3AS}. 

\textit{General ideas.} A different way to construct the vector field $v$ is to adopt the strategy introduced in \cite{allaire2004JCP} based of the computation of the shape derivatives \cite{allaire2007book, vandijk2013SMO}.
In this approach, the user must suitably define a shape functional $J(\Omega)$ which maps any feasible subset of $\R^3$ to a real number. The subset $\Omega$ represents the solid surface one can possibly create, and $J(\Omega)$ somehow measures the ``cost'' of that object. 
The functional $J$ is therefore defined in such a way as to penalize the undesired features of the surface. In the case of interest, the feature to penalize is the presence of overhangs and the computational strategy consists of finding a suitable scalar velocity field $v$ which drives the evolution of the shape $\Omega_t$ (using the level-set method) in such a way that the function $t\mapsto J(\Omega_t)$ is decreasing. 
At the end of the shape optimization procedure we are hopefully left with a fully printable object $\Omega_*$, possibly different from the original one and then, as we did in Section \ref{rocchi} we just need to remove the excess part $\Omega_*\backslash\Omega_0$.  

Roughly speaking, the idea is explained as follows (see \cite{allaire2007book, allaire2004JCP} for a complete discussion and mathematical details):
let $\Gw$, with $|\Gw|<1$, be a given vector field and let $O$ a given bounded domain of $\R^3$. 
Interpreting $\Gw$ as a displacement of the domain $O$, we get the new domain $O^{\Gw}$ defined by
$$
O^{\Gw}:=\left\{(x,y,z)+\Gw(x,y,z) \ : \ (x,y,z)\in O  \right\}.
$$
By definition, if the functional $J$ is shape differentiable at $O$ in the direction $\Gw$ the following expansion holds 
\begin{equation}\label{sensitivityofJ}
	J(O^{\Gw})=J(O)+J^\prime(O)(\Gw)+o(\Gw)
\end{equation} 
where the function $J^\prime(O)(\Gw)$ denotes the shape derivative of $J$ in the direction $\Gw$.

Whenever the expression of the shape derivative falls in the general form \begin{equation}\label{shapederivativegeneralform}
	J^\prime(O)(\Gw)=\int_{\partial O}\pi(x,y,z)\ \Gw(x,y,z)\cdot \Gn(x,y,z) \ ds
\end{equation} 
for some scalar function $\pi$, we can easily get some useful information about the corresponding shape optimization problem. 
Indeed, by choosing as displacement direction $\Gw$ a vector field defined as
$$
\Gw=-\pi\Gn,
$$
we immediately get 
\begin{equation*}
	J^\prime(O)(\Gw)=-\int_{\partial O}\pi^2(x,y,z)\  ds<0.
\end{equation*}
By \eqref{sensitivityofJ}, we deduce that any displacement in the direction $\Gw=-\pi\Gn$ lead to a deformation of the domain which \emph{lowers} the value of $J$.

\medskip

\emph{Example.} It is useful to note that the shape derivative can be explicitly computed in some cases. If we consider the shape functional $$
J(O)=\int_O \pi(x,y,z)dx dy dz
$$ 
for some scalar function $\pi$ (in the particular case $\pi\equiv 1$, $J$ gives the volume of the domain), the shape derivative is exactly \eqref{shapederivativegeneralform}; see \cite[Prop.\ 6.22]{allaire2007book} for the detailed derivation. $\qed$

\medskip

Now, recalling how the level-set method describes the front evolution $t\to\Omega_t$ (see Eq.\ \eqref{eikonaleevolutiva}), we get that transporting $\u$ by the dynamics
\begin{equation}\label{eikevallaire}
	\partial\u_t-\pi |\nabla \u|=0
\end{equation}
is equivalent to move the boundary $\partial\Omega_t$ (i.e.\ the zero level-set of $\u(t)$) along the descent gradient direction $-J^\prime(\Omega_t)$, thus evolving towards a lower cost shape.

\medskip

\emph{Definition of $J$.}
In \cite{allaire2004JCP} the authors propose a functional which is proven to be effective in minimizing overhangs.
The idea is the following: first, let us denote by $H$ the height of the object. Given $h\in[0,H]$, we assume that the surface was already built until height $h$, and now the layer at height $h$ is being processed.
Let us consider now that the only force acting on the surface $\Sigma$ of the object $\Omega$ is the gravity, which pushes down the object slightly deforming its shape. 
The elastic displacement $\mathbf u_h(x,y,z)\in\R^3$ due to this deformation can be computed solving  the well known equation
\begin{equation}\label{displacement-h}
	\left\{
	\begin{array}{ll}
		-\nabla \cdot \sigma(\mathbf u_h)=\mathbf g,  & \quad (x,y,z)\in\Omega  \\
		\mathbf u_h=\mathbf 0, & \quad (x,y,z)\in\Sigma_{\downarrow} \\
		\sigma(\mathbf u_h)\Gn=\mathbf 0, & \quad (x,y,z)\in\Sigma\backslash\Sigma_{\downarrow} 
	\end{array}
	\right.
\end{equation}
where 
$\Sigma_{\downarrow}$ is the contact region between $\Omega$ and the build table, 
$\mathbf g(x,y,z)$ is the body force (in our case, the gravity),
$\sigma(\mathbf u)\in\R^{3\times 3}$ is the matrix defined as
$$
\sigma(\mathbf u):=\lambda(\nabla\cdot\mathbf u)\mathbf I+\mu \left(\nabla\mathbf u + (\nabla\mathbf u)^T\right),
$$
$\lambda$ and $\mu$ are the Lam\'e coefficients of the material, and $\mathbf I$ is the $3\times 3$ unity matrix. Moreover, we have defined
$$
\begin{array}{l}
\nabla\cdot\mathbf u:=u^1_x+u^2_y+u^3_z
\\ \phantom{x} \\
\nabla\mathbf u:=
\left(\begin{array}{ccc}
u^1_x & u^1_y & u^1_z \\ 
u^2_x & u^2_y & u^2_z \\ 
u^3_x & u^3_y & u^3_z \\ 
\end{array}\right) 
\\ \phantom{x} \\
\nabla\cdot\sigma:=
\left(\begin{array}{c}
\sigma^{11}_x + \sigma^{12}_y + \sigma^{13}_z\\ 
\sigma^{21}_x + \sigma^{22}_y + \sigma^{23}_z \\ 
\sigma^{31}_x + \sigma^{32}_y + \sigma^{33}_z \\ 
\end{array}\right).
\end{array}
$$

Given the displacement $\mathbf u_h$, it is easy to compute the \emph{compliance}, i.e.\ the work done by the forces acting on the object. 
In our case we have
\begin{equation}\label{compliance-h}
	c_h=\int_\Sigma \mathbf g \cdot \mathbf u_h ds.
\end{equation}
Minimizing the compliance corresponds to maximize the rigidity of the object with respect to the forces acting on it. 
In our case this rigidity is automatically translated in a minimization of the overhangs, since deformations are greatest precisely along the protrusions. 

Since overhangs should be minimized over the entire surface from bottom to top, the final cost functional $J$ is defined by summing the compliance over all layers
\begin{equation}\label{Jallaire}
	J(\Omega)=\int_0^H \! c_h \ dh.
\end{equation}

Surprising enough, the shape derivative of $J$ can be computed explicitly and falls in the general form \eqref{shapederivativegeneralform}, with a suitable function $\pi$ which depends on $\mathbf u_h$, see \cite{allaire2017JCP} for details.

In conclusions, one can find an overhang-free surface by evolving the initial surface $\Sigma_0$ (enclosing the domain $\Omega_0$) in the direction $-\pi(\mathbf u_h)$ by means of the level-set equation \eqref{eikevallaire}, considering that at each time $t$ the vector $\mathbf u_h$ must be recomputed solving \eqref{displacement-h} over the time-variable domain $\Omega_t$ (enclosed in $\Sigma_t$).

\section{Building object-dependent infill structures}\label{sec:tanzilli}
Beside difficulties related to 3D-printing overhanging parts, another issue arises when one has to manufacture an object: what are \emph{inside} the surface?
In principle, the ideal choice would be to print fully solid objects. In this way, the object has maximal rigidity and the internal overhanging parts  are resolved (e.g., the top half of the surface of a sphere, which should lie on a support inside the sphere in order not to fall down).
Unfortunately, printing fully solid objects is often not convenient because of the large quantity of material to be used. 
Shape optimization tools can give the optimal way to hollow out the object, reducing the overall material volume and keeping at the same time the desired rigidity and printable features. The problem reduces to finding the optimal inner structure supporting the whole object from the inside.

\medskip

\begin{figure}[h!]
	\begin{center}
		\includegraphics[width=3.98cm]{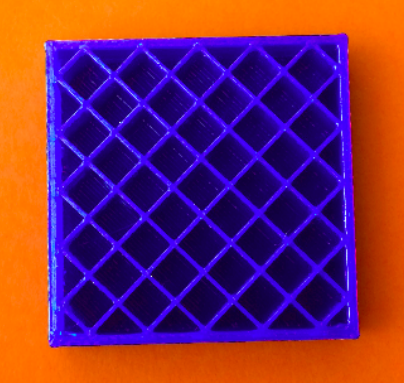}
		\qquad\qquad
		\includegraphics[width=4cm]{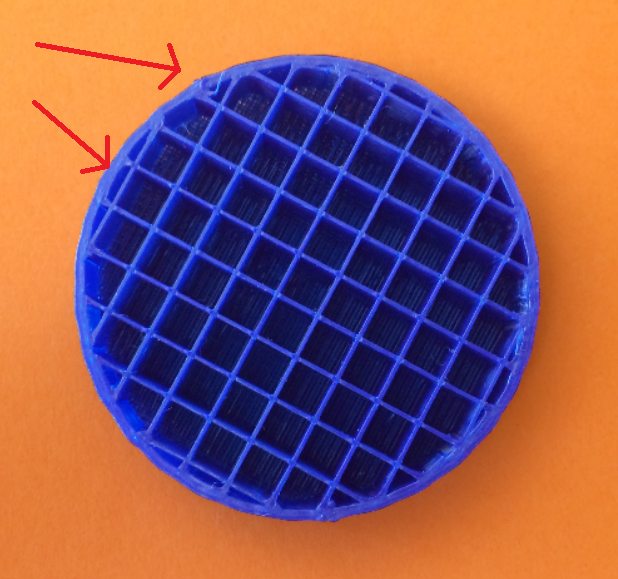}
	\end{center}
	\caption{Predefined square pattern for infill structure. In some parts the infill lines are very close to the boundary (external surface) of the object and/or the lines are  very close to each other. As a consequence, the 3D printer hardly resolves the pattern, the extruder wastes time running for very short distance and the material already deposed melts again.}
	\label{fig:unoptiamlinfill}
\end{figure}

Commercial software typically creates infill structures with a specific pattern (e.g., squares, honeycomb, etc.) which are independent of the object under consideration. 
This is an easy and fast solution, but it is clearly nonoptimal is some cases, see, e.g., Fig.\ \ref{fig:unoptiamlinfill}.

In the following we propose a method inspired by \cite{dhanik2010IJAMT, kimmel1993CAD} to create an object-dependent optimal infill structure specifically designed to follow the contours of the layer to be printed. 
This approach minimizes the changes of direction of the extruder and fix the problem shown in Fig.\ \ref{fig:unoptiamlinfill}, avoiding to create holes of different size.

The method is again addressed to 3D printers with FDM technology, in which the object is created layer by layer starting from the lowest one, and it is again based on a front propagation problem and the level-set method, see Sect.\ \ref{sec:LSmethod}.
Differently from what we did before, here the level-set problem will be set in 2D (with the level-set function $\u$ defined in $\R^+\times \R^2$).

The idea is described in Fig.\ \ref{fig:infill_pipeline} and can be summarized as follows: 
\begin{figure}[h!]
	\begin{center}
		(a)\ \includegraphics[width=2.6cm]{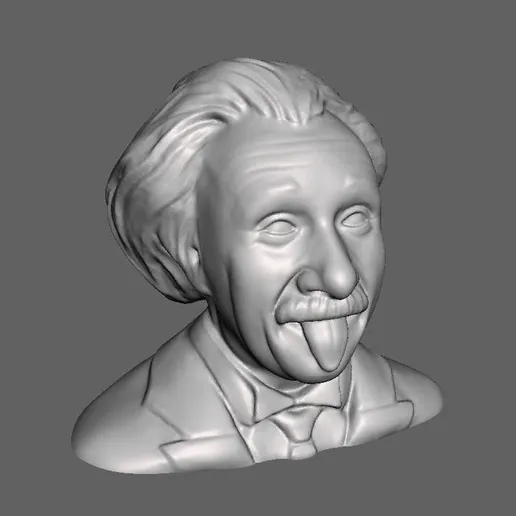}\quad
		(b)\ \includegraphics[width=4.4cm]{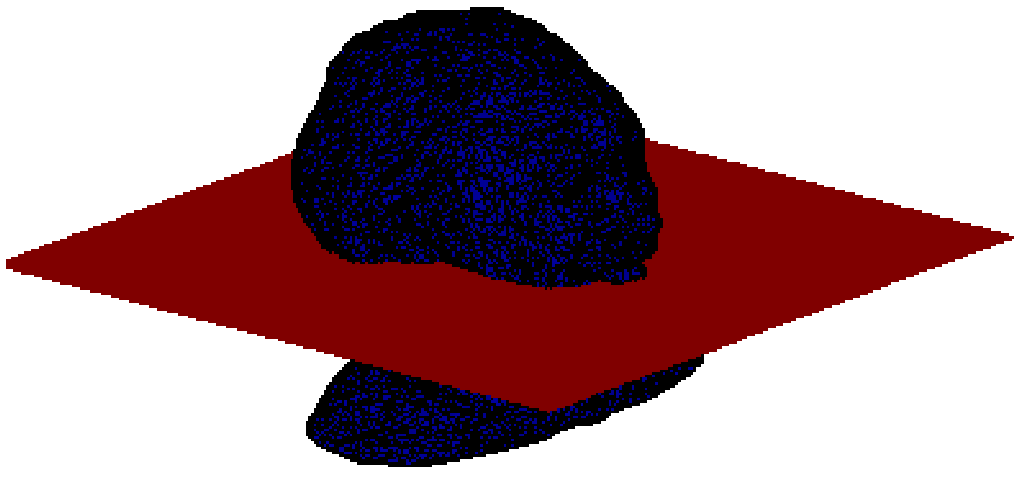}\quad
		(c)\ \includegraphics[width=2.8cm]{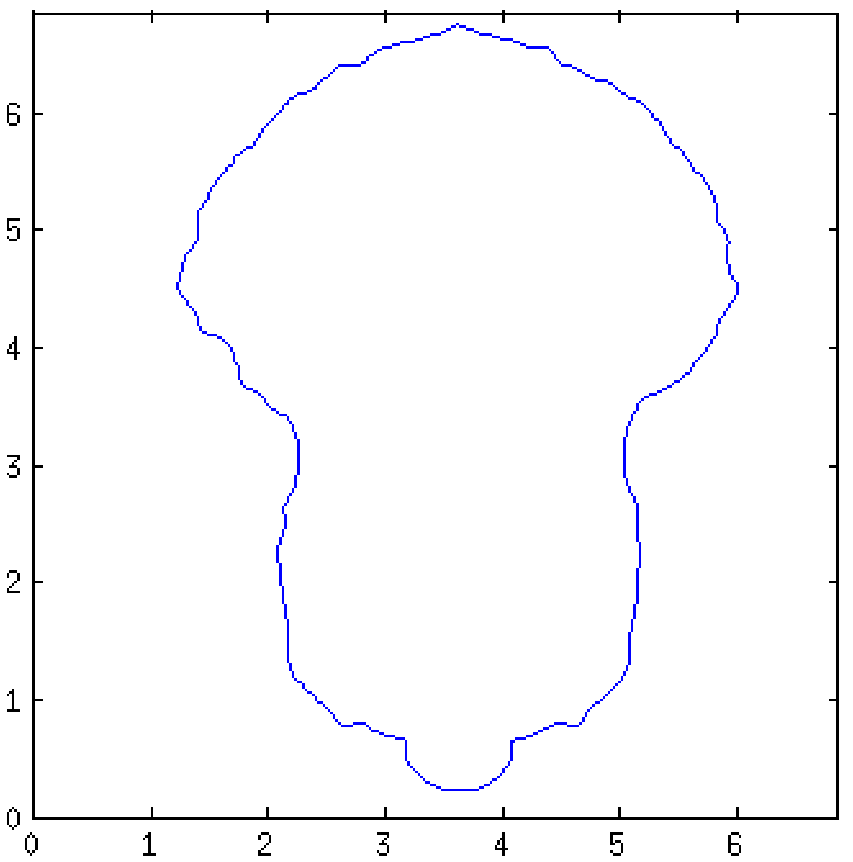}\\
		(d)\ \includegraphics[width=3.5cm]{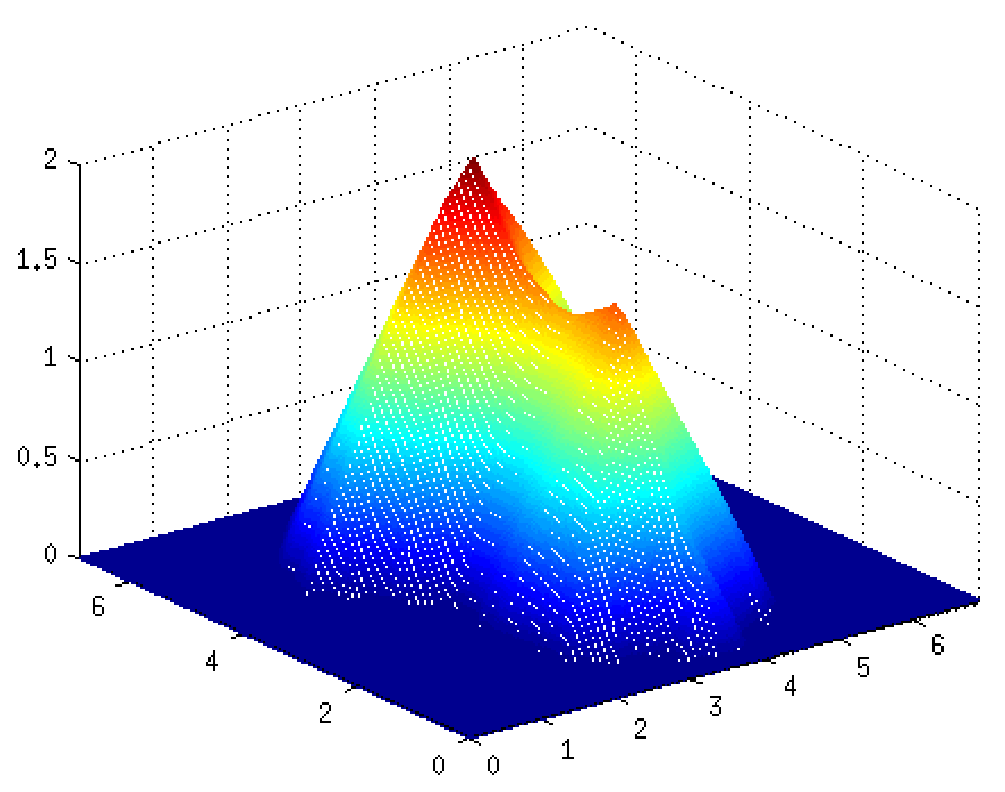}\quad
		(e)\ \includegraphics[width=3.5cm]{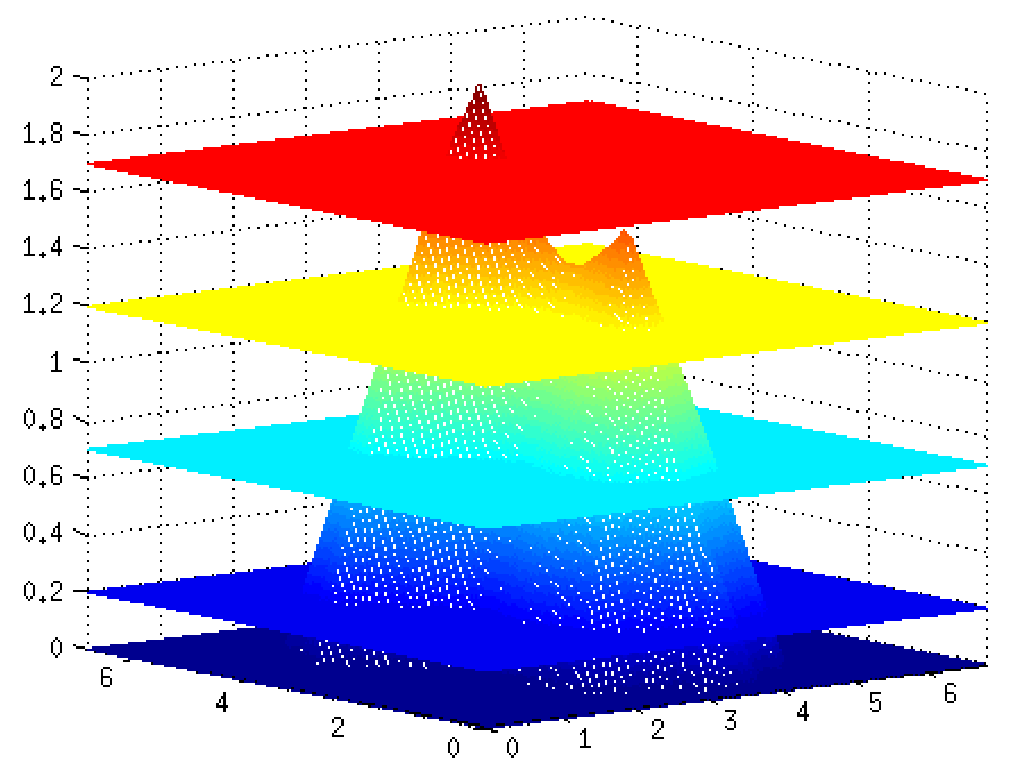}\quad
		(f)\ \includegraphics[width=2.9cm]{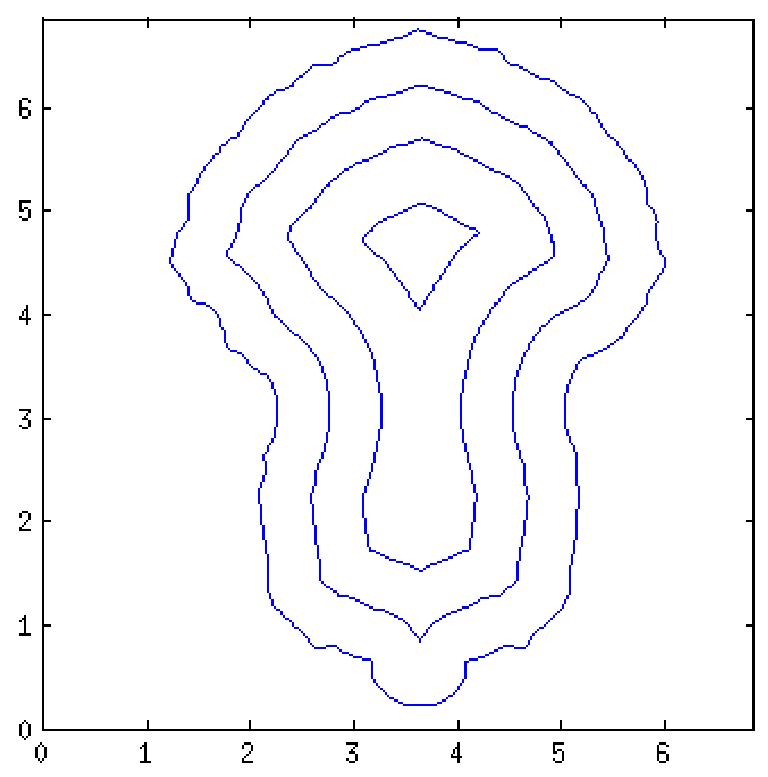}
	\end{center}
	\caption{
		(a) Original 3D object to be printed. 
		(b) Object cut at height $h$ (layer). 
		(c) Contour $\Gamma_h$ of the layer.
		(d) Distance function from the contour $\Sigma_h$, solution to the eikonal equation.
		(e) level-sets of the distance function.
		(f) level-sets as infill structure of the layer.
	}
	\label{fig:infill_pipeline}
\end{figure}
\begin{itemize}
	\item [(a)]\ Start from the surface to be printed, empty inside, and given in terms of facets (little triangles which cover the whole surface without leaving holes) with oriented vertices to distinguish the inner and the outer space.
	\item [(b)]\ Cut the surface with a horizontal plane at height $h$. This gives the contour of the object and the layer to be printed at the current step.
	\item [(c)]\ Denote the layer as $\Gamma_h$. This is the initial front (1D, embedded in $\R^2$) we want to evolve through the level-set equation in the inner normal direction $-\mathbf N$ until it shrinks to a single point. 
	\item [(d)]\ Solve the isotropic stationary eikonal equation $|\nabla T_h(x,y)|=1$ inside $\Gamma_h$, with $T_h(x,y)=0$ for $(x,y)\in\Gamma_h$. This gives the signed distance function from $\Gamma_h$ \cite{sethianbook}.
	\item [(e)]\ Cut the solution at different heights, the number of levels depends on how fine we want the infill structure.
	\item [(f)]\ The level-sets of the solution correspond to the desired infill structure, which, by construction, follows the boundary of the object.
\end{itemize}
The procedure is then repeated for each layer until the top of the object is reached. 

\medskip

\begin{figure}[h!]
	\begin{center}
		\begin{tabular}{ccc}
			\includegraphics[width=3.9cm]{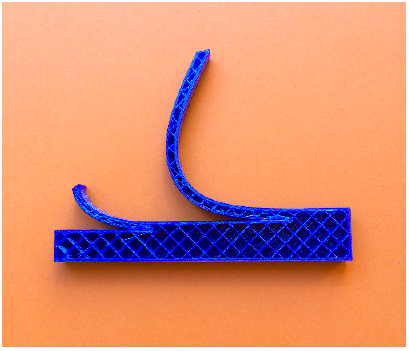}&
			\includegraphics[width=4.1cm]{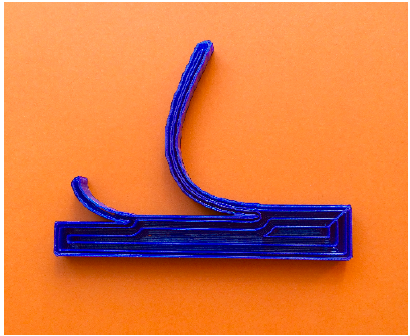}&
			\includegraphics[width=2.5cm]{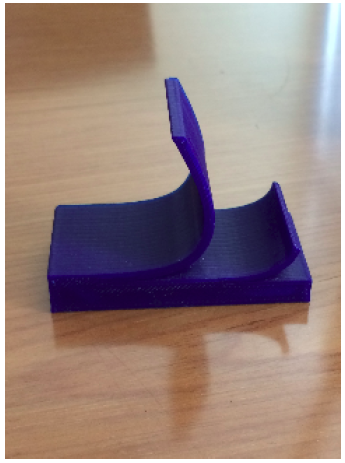}\\
			\footnotesize (a) & \footnotesize (b) & \footnotesize (c)  
		\end{tabular}
	\end{center}
	\caption{A 3D-printed  phone holder. (a) square infill, (b) proposed infill, (c) final object.}
	\label{fig:portacell}
\end{figure}

In Fig.\ \ref{fig:portacell} we show a real application of the method described above, 3D-printing a phone holder. It is basically a parallelepiped with two thin curved branches. 

We have compared the eikonal-based infill with the one obtained by a predefined square pattern. The G-code (see Appendix B) was directly coded, without resorting to a commercial slicer.
It is clear that with such thin branches the new method can have an advantage over a general-purpose one, derived by following the original shape of the object's boundaries. 
Therefore we expect a mild advantage in terms of extruded material and a more important advantage in terms of printing time. 
In Table \ref{tab:cfrTanzilli} we summarized the results which indeed confirm the expectations.
\begin{table}[h!]
	\begin{center}
		\begin{tabular}{l|l|l|r|}
			&  \ SQUARE \ & \ PROPOSED \ & \ difference \ \\ \hline \hline
			time          &  18 m 30 s  & 11 m 24 s    & -38\% \\ \hline
			material \  &  1559 mm  & 1420 mm    & -9\% \\ \hline
		\end{tabular}
	\end{center}
	\caption{Comparison between the square infill and the proposed one in terms of printing time and extruded material. 3D printer was Delta WASP 2040.}
	\label{tab:cfrTanzilli}
\end{table}

\section{Conclusions}
In this chapter we have examined various mathematical models and  techniques that can be applied to 3D reconstruction and 3D printing and that allow to deal with both problems in a unified manner. In particular, we have seen that nonlinear PDEs appear for the modeling  of complex computer vision problems ranging from the classical case of a single image with orthographic projection under the Lambertian assumption to more general surfaces and to the multi-view problem. The starting point in the simplest case is the eikonal equation but the difficulty rapidly increases to deal with systems of nonlinear PDEs. 
 
As for the 3D printing, we have examined some technical problems that can be solved via the level-set method for front propagation and its generalization in the context of shape optimization. Again, similar equations appear and allow to solve at least two important problems for 3D printing: the detection and resolution of overhangs, and the optimization of the deposition path to actually build the object.
 
We hope that this chapter will attract the attention of young researchers to this new area, which still presents many open problems as well as an increasing importance for industrial manufacturing of small and even large objects.

\appendix
\section*{Appendix A: The STL format} \label{appendixSTL}
STL (STereo Lithography interface format or Standard Triangulation Language) is the most common file format used to store object data which have to be 3D-printed.
It comes in two flavors, ASCII or binary. The latter requires less space to be stored and it can be easily created from the former via free software. 

Objects are determined by means of their surface, which must be closed, so to be watertight. 
It is also mandatory that the ``interior'' and the ``exterior'' of the solid are always defined. 
The surface of the solid to be printed must be tessellated by means of a triangulation. 
Each triangle, commonly called \emph{facet}, is uniquely identified by the $(x,y,z)$ coordinates of the three vertices $(\mathbf V^1,\mathbf V^2,\mathbf V^3)$ and by its unit exterior normal $(N_x, N_y, N_z)$. The total number of data for each facet is 12. 
Moreover,
\begin{enumerate}
	\item each triangle must share two vertices with every neighboring triangle. In other words, a vertex of one triangle cannot lie on the side of another triangle;    
	\item all vertex coordinates must be strictly positive numbers.  The STL file format does not contain any scale information, the coordinates are in arbitrary units. Actual units (mm, cm, ...) will be chosen in the printing process;
	\item each facet is part of the boundary between the interior and the exterior of the object. The orientation of the facets is specified redundantly in two ways. First, the direction of the normal is outward. Second, the vertices are listed in counterclockwise order when looking at the object from the outside (right-hand rule).
\end{enumerate}
The actual format of the STL file is given in the following.
\lstset{basicstyle=\ttfamily, showspaces=false, backgroundcolor=\color{coloresfondo}}
\begin{lstlisting}
solid name-of-the-solid
  facet normal nx ny nz
    outer loop
      vertex V1x V1y V1z
      vertex V2x V2y V2z
      vertex V3x V3y V3z
    endloop
  endfacet
  facet normal nx ny nz
    outer loop
      vertex V1x V1y V1z
      vertex V2x V2y V2z
      vertex V3x V3y V3z
    endloop
  endfacet
  [...]
endsolid name-of-the-solid
\end{lstlisting}
Values are \emph{float} numbers and indentation is made by two blanks (no tab). Unit normal direction can be simply computed by
$$
\mathbf N=\frac{(\mathbf V^2-\mathbf V^1)\times (\mathbf V^3-\mathbf V^2)}{|(\mathbf V^2-\mathbf V^1)\times (\mathbf V^3-\mathbf V^2)|}.
$$

\section*{Appendix B: The G-code format}
The STL file described above is far from being processed by a 3D printer. Another step is needed in order to transform the set of facets composing the surface of the object in something which can be understood by the machine.

The G-code is a programming language for CNC (Computer Numerical Control) machines and 3D printers. 
In the case of 3D-printers, it is used to tell the machine the exact path the extruder should follow, how fast to move and how much material should be deposed.
Putting the infill problem aside for a moment, the extruder's path is obtained by slicing the triangulated surface by horizontal planes and taking the intersection between the facets and each plane, see Fig.\ \ref{fig:infill_pipeline}(a-c).
The resulting curve is then approximated by a sequence of points. 
The machine will pass through these points in the given order, extruding the material. 

The core of a G-code file sounds like this:
\lstset{basicstyle=\ttfamily, showspaces=false, backgroundcolor=\color{coloresfondo}}
\begin{lstlisting}
G1 F# X# Y# Z# E#
G1 F# X# Y# Z# E#
G1 F# X# Y# Z# E#
[...]
\end{lstlisting}
where $\sharp$ is a \emph{float} number. 
G1 means that the extruder has to move along a straight line (circular patterns are also possible with G2 and G3), F is the speed of the extruder, the following triple X,Y,Z represents the coordinates of the point to be reached, in this step, from the current position of the extruder. Finally, E indicates how much material has to be extruded up to that line of code.

Many other commands exist and are used to start/stop the machine, set the units of measure, set the temperature of the nozzle/bed, set fan speed, etc. 

\begin{acknowledgement}
The authors want to acknowledge the contribution of F. Tanzilli for the experiments presented in Section \ref{sec:tanzilli}.
The authors are members of the INdAM Research Group GNCS. 
This work was carried out within the INdAM-GNCS project "Metodi numerici per l'imaging: dal 2D al 3D", Code CUP\_E55F22000270001.
\end{acknowledgement}

\bibliographystyle{plain}
\bibliography{biblio3DV,biblio3DP}

\end{document}